\documentclass{article}
\usepackage{deeprobot_paper,times}

\usepackage{amsmath,amsfonts,bm}

\def\eqref#1{equation~\ref{#1}}
\def\1{\bm{1}}

\DeclareMathAlphabet{\mathsfit}{\encodingdefault}{\sfdefault}{m}{sl}
\SetMathAlphabet{\mathsfit}{bold}{\encodingdefault}{\sfdefault}{bx}{n}

\usepackage{xcolor}
\usepackage{booktabs}
\usepackage{graphicx}
\usepackage{array}
\usepackage{colortbl}

\newcommand{\method}{DR-WM}

\newcommand{\protocol}{\paragraph{Protocol.}}
\newcommand{\datasplit}{\paragraph{Data split.}}
\newcommand{\baselines}{\paragraph{Baselines.}}
\newcommand{\metrics}{\paragraph{Metrics.}}

\usepackage{float}
\usepackage{enumitem}
\usepackage{etoolbox}
\usepackage{hyperref}
\usepackage{url}

\title{From Pixels to Executable Physics:\\
Behavior-Grounded World Models for Embodied Action Consequence Prediction}

\author{%
  Ruo Chen\textsuperscript{1,2,*}\quad
  Jiahua Pang\textsuperscript{1,3,*}\quad
  Yingjie NIU\textsuperscript{1,\ensuremath{\dagger},\ensuremath{\ddagger}}\quad
  Hongyi Lin\textsuperscript{4,5}\\[0.25em]
  Jun Tian\textsuperscript{1,6}\quad
  Yang Zhu\textsuperscript{2,\ensuremath{\dagger}}\quad
  Chao Li\textsuperscript{1,\ensuremath{\dagger}}%
}
\affiliation{\small
  \textsuperscript{1}DEEP Robotics\quad
  \textsuperscript{2}Zhejiang University\quad
  \textsuperscript{3}Beijing Institute of Technology\\[0.15em]
  \textsuperscript{4}Massachusetts Institute of Technology\quad
  \textsuperscript{5}Tsinghua University\\[0.15em]
  \textsuperscript{6}University of Illinois at Urbana-Champaign\par
  \vspace{0.3em}%
  {\footnotesize\sffamily\color{DeepMuted}%
    \textsuperscript{*}Equal contribution\quad
    \textsuperscript{\ensuremath{\ddagger}}Project leader\par
    \textsuperscript{\ensuremath{\dagger}}Corresponding authors:
    Yingjie NIU:
    \href{mailto:niu.yingjie@g.sp.m.is.nagoya-u.ac.jp}{niu.yingjie@g.sp.m.is.nagoya-u.ac.jp}\\[0.15em]
    Yang Zhu: \href{mailto:zhuyang88@zju.edu.cn}{zhuyang88@zju.edu.cn}\quad
    Chao Li: \href{mailto:lichao@deeprobotics.cn}{lichao@deeprobotics.cn}\\[0.15em]
    Project: \url{https://developer.deeprobotics.cn/research/articles/dr-wm/index.html}}%
}
\paperdate{}

\begin{document}
\makebrandheader
\begin{deeprobotfrontmatter}
  \maketitle
  \begin{abstract}
In outdoor environments, varying terrain and contact conditions can cause identical action commands to produce different motion outcomes for quadruped robots. 
Predicting these outcomes is essential for reliable navigation and informed path selection. 
Existing video world models can generate visually plausible futures, yet often struggle to capture robot dynamics and terrain interactions, leading to overly optimistic predictions of action outcomes. 
We present \method{}, a physically consistent world model that incorporates robot–terrain contact dynamics to predict the consequences of quadruped actions. 
Starting from an initial RGB image, our method reconstructs scene geometry and integrates dense, dimensionless priors over terrain contact properties, calibrated using real-world action–response data. The
 resulting terrain model supports the simulation of physics with spatially varying contact properties such as friction. 
Given the initial state of the robot and candidate action sequences, we simulate its future motion and use the resulting camera trajectories and rendered geometry to guide the training of a video world model.
The generated future observations are then used to update the terrain reconstruction, enabling iterative model rollouts without additional real-world images. 
Extensive experiments demonstrate that \method{} substantially outperforms geometry-only baselines and baselines with uniform physical parameters on key physical metrics characterizing action outcomes, while achieving the best performance among the evaluated baselines. 
These physically grounded predictions support more informed action selection: across 90 challenging scenarios, an 8B vision–language model augmented with \method{} outperforms an unassisted 32B vision–language model on five of six evaluation metrics.

% Robots increasingly navigate complex environments, yet predicting action
% consequences on unfamiliar terrain remains challenging. Changing contact
% conditions can turn steady progress into slipping, making real-world evaluation
% costly and potentially hazardous.
% World models can assess candidate actions before execution.
% However, visually plausible futures do not ensure that predicted motion
% reflects robot dynamics and terrain contact. We introduce \method{}, a
% behavior-grounded world model for robot-terrain interaction. We combine
% geometry reconstructed from the initial RGB image with dense, dimensionless
% physical priors calibrated using real action--response data. Using this
% executable terrain, we simulate motion from the initial robot state and action
% sequence. The resulting camera trajectories and rendered geometry guide our
% trained video model, whose generated observations update reconstruction for
% continued rollout without further real images. On held-out test scenes,
% \method{} outperforms geometry-only and uniform-physics alternatives on key
% physical metrics characterizing robot action consequences and leads all reported automatic
% and reference-aligned video metrics. By providing physically plausible future observations,
% \method{} supports more informed action selection: across 90 challenging scenes,
% an 8B vision-language model with \method{} outperforms an unassisted 32B model
% on five of six metrics.
\end{abstract}

\end{deeprobotfrontmatter}

\begingroup
\setlength{\intextsep}{4pt}
\setlength{\abovecaptionskip}{4pt}
\setlength{\belowcaptionskip}{0pt}
\begin{figure}[!htbp]
\centering
\includegraphics[
  page=1,
  width=\linewidth
]{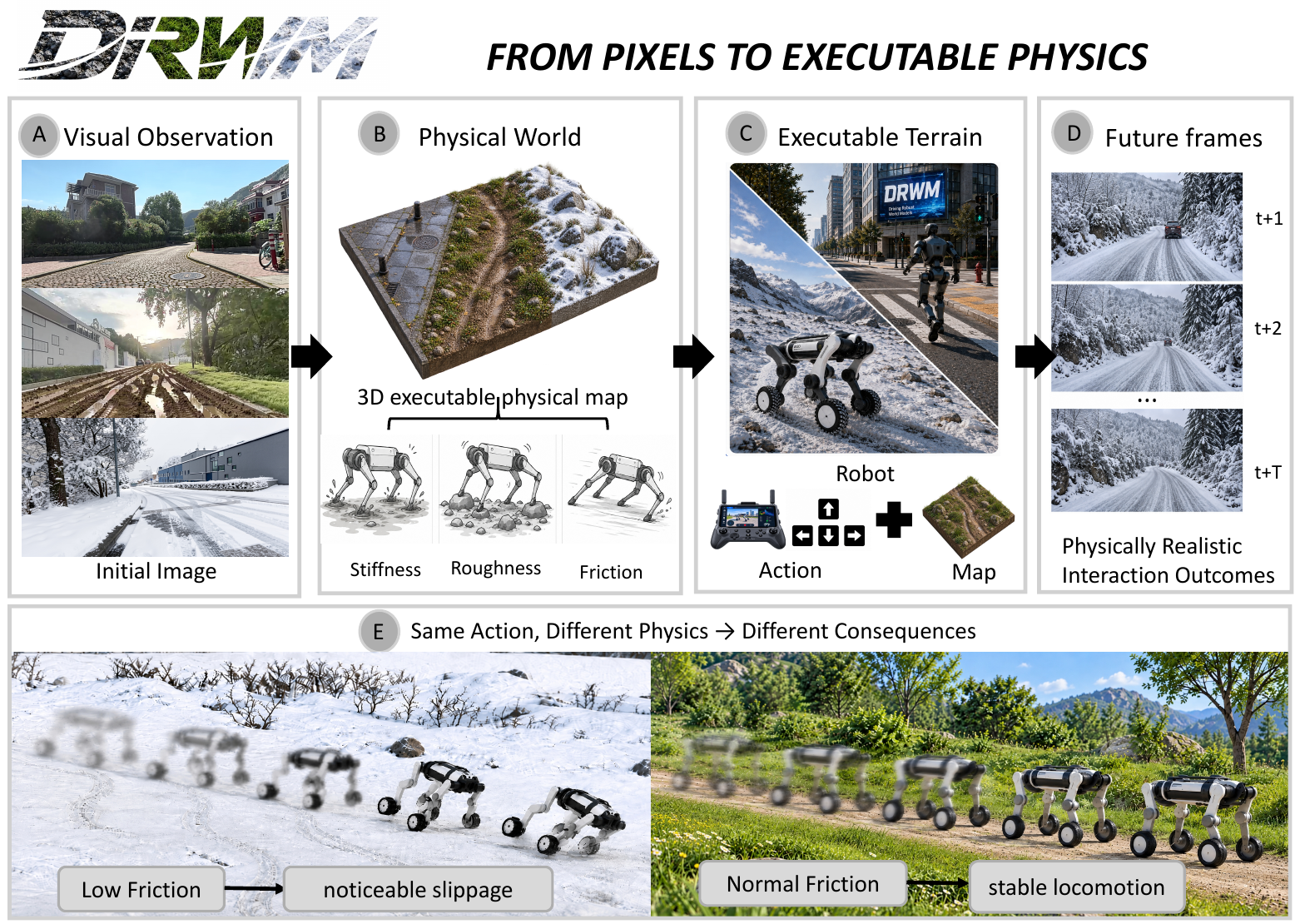}
\caption{Overview of \method. The model constructs an executable terrain representation from visual observations, then simulates terrain dynamics to predict future observations and terrain-dependent action outcomes.}
\label{fig:method-overview}
\end{figure}
\endgroup

\section{Introduction}
\label{sec:introduction}
Outdoor quadruped navigation depends not only on scene semantics and geometric traversability but also on how terrain contact affects robot motion. 
As embodied navigation moves from structured laboratory settings to complex outdoor environments, robots encounter spatially varying friction, support, and contact conditions. 
Under the same low-level controller, an identical nominal motion command may produce steady progress, slipping, attitude disturbances, or even instability~\citep{chen2024identifyingterrainphysicalparameters}. 
Reliable outdoor navigation therefore requires both understanding what is in the scene and predicting what will happen when an action is executed on the current terrain.

A natural approach is to predict candidate action outcomes before execution, rule out high-risk behaviors, and decide whether to proceed or adjust motion. However, real-world robot trials are costly and time-consuming. 
World models offer a promising alternative, enabling policies or planners to compare candidate actions through visual predictions of their outcomes~\citep{ha2018worldmodels,hafner2019dreamer,hafner2019learninglatentdynamicsplanning}.

However, a visually plausible future is not necessarily physically realizable. 
Existing video world models typically generate future observations in two ways. 
One class conditions on predefined camera trajectories to synthesize viewpoint changes and scene motion. 
This assumes the ground-truth camera trajectory and cannot determine whether it is physically feasible on the current terrain. 
Another class predicts video directly from actions, but most current applications focus on game generation and remain far removed from real-world settings. 
Scene geometry alone is also insufficient: geometrically similar surfaces can produce markedly different robot responses due to different contact properties. Consequently, existing models may depict smooth forward motion even when the same action would cause severe slipping or instability in reality.

In this research we focus on physical consistency between robot motion and visual predictions: candidate actions and spatially varying terrain contact jointly determine the robot state trajectory, from which camera motion is derived. 
This poses two key challenges. 
First, a single RGB image provides only uncertain priors over terrain properties~\citep{chen2024identifyingterrainphysicalparameters}. 
These priors require calibration using real robot action–response data to obtain contact parameters that reproduce robot motion. 
Second, as the robot moves beyond the initial field of view, the model must complete unobserved regions and estimate their contact properties. Although video models can complete visual content using scene priors and coarse 3D information~\citep{wan2025,zhao2025spatiavideogenerationupdatable}, long-horizon motion predictions may still deviate from real outcomes without behavior-grounded physical representations of these regions.

To address these challenges, we propose \method{}, a physically consistent world model for robot–terrain interaction (Figure~\ref{fig:method-overview}). 
Given an initial RGB image, the robot’s initial state, and candidate action sequences, \method{} jointly predicts future robot motion and the corresponding visual observations. 
Our core idea is to model physical evolution and visual completion separately while coupling them through a shared robot state trajectory: contact dynamics determine how the robot moves, and the video model predicts what it will see as a result. Unlike methods that guide video generation with coarse simulation~\citep{li2025wonderplaydynamic3dscene}, \method{} predicts motion using a terrain contact model calibrated against real robot behavior.

\begin{figure}[!htbp]
\centering
\includegraphics[page=1,width=\linewidth]{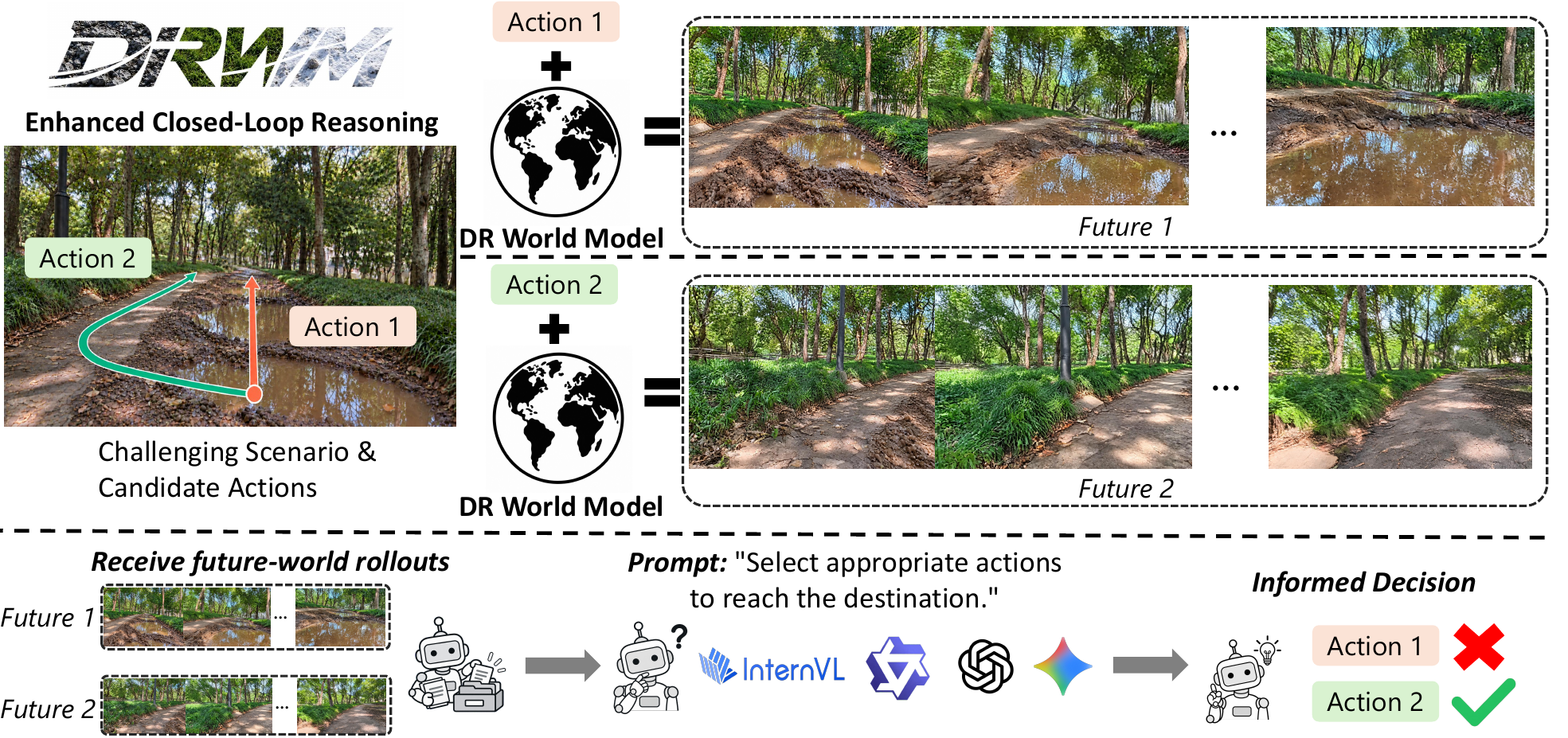}
\caption{VLM action-selection pipeline. For each candidate action, \method{} predicts future observations and iteratively feeds them back into scene reconstruction and physics simulation. The VLM compares the resulting rollouts and selects the safer action. In this example, it chooses the detour (Action 2) over traversing the muddy terrain (Action 1); see Section~\ref{sec:exp-vlm-decision}.
}
\label{fig:vlm-action-selection}
\end{figure}

Specifically, \method{} integrates terrain modeling, physics-guided video generation, and iterative scene updates. 
First, we reconstruct scene geometry from visual observations and train PhyMap to predict dense, normalized priors for contact properties. 
We then map these priors to effective contact parameters using real-world robot action–response data. This calibration mapping is learned exclusively from the training data and remains fixed at test time. 
Second, for each candidate action, we simulate the resulting robot motion using the calibrated contact parameters and derive the corresponding camera trajectory from the predicted base poses~\citep{genesis2024}. 
Conditioned on this trajectory and coarse geometric renderings, the video model generates future frames. Finally, we use the generated frames to estimate the geometry and contact properties of previously unobserved regions. 
By propagating the robot state across chunks and incrementally extending the terrain representation, \method{} supports long-horizon rollouts without requiring additional real-world images. 
A decision model then evaluates the candidate actions using these predicted observations and selects the action expected to produce more stable, lower-risk motion (Figure~\ref{fig:vlm-action-selection}). Experiments show that the behavior-calibrated, spatially varying contact model substantially outperforms geometry-only and uniform-physics baselines on key motion-prediction metrics. 
\method{} further outperforms every competing baseline on all reported video metrics. 
Across 90 challenging action-selection scenarios, an 8B VLM augmented with \method{} outperforms a standalone 32B VLM on five of the six evaluation metrics.
% \begin{itemize}[nosep,topsep=0pt,partopsep=0pt,beginpenalty=10000,midpenalty=0]
% \interlinepenalty=10000
% \item \textbf{Behavior-calibrated spatial terrain modeling.}
% We use real robot action-response data to map dense visual priors over terrain properties to effective simulator parameters that reproduce robot motion, enabling an executable terrain model with spatially varying contact conditions.
% \item \textbf{Physics-grounded visual prediction.}
% We first predict future robot motion through robot-terrain contact dynamics and then condition video generation on the camera trajectory and coarse geometry derived from the same simulated state trajectory. This design ties the predicted visual motion to the physical outcome of the candidate action.
% \item \textbf{State-continuous rollouts for action selection.}
% We develop an iterative rollout process that preserves robot state across segments and expands the internal terrain representation using predicted frames. We further show that these physically grounded rollouts improve motion prediction, video generation, and VLM-based candidate action selection.
% \end{itemize}
Our main contributions are as follows:
\begin{enumerate}[label=\textbf{(\arabic*)},wide=0pt,leftmargin=0pt,listparindent=0pt,itemsep=3pt,parsep=0pt,topsep=3pt]

\item \textbf{Physical World Model for Robot Navigation.} 
% We propose a unified framework for physically plausible action consequence prediction in navigation through terrain understanding, behavioral simulation, and video generation. 
We propose \method{}, a unified framework for terrain-dependent action consequence prediction. It integrates terrain understanding, behavior-calibrated physics simulation, and a video model tailored to robot motion prediction.
\item \textbf{Large-Scale Multimodal Physical Dataset.}
We introduce \textbf{Physcene}, comprising two distinct subsets.
\textbf{Physcene--Image} contains 2.58M images with dense physical priors.
\textbf{Physcene--Robot} provides over 100K dedicated robot clips with visual observations, robot states, and control signals.
\item \textbf{Physically Plausible Simulation.} 
\method{}'s autoregressive reasoning achieves  the best performance on \textbf{Physcene--Robot} evaluations, making world models practically useful for real-world robot navigation.

\end{enumerate}

\section{Related Work}
\label{sec:related}

\subsection{Visual and Action-Conditioned World Models}

Visual world models have evolved from latent prediction to controllable video generation.
Early approaches predict future observations for
action evaluation~\citep{finn2017deepvisualforesightplanning,
hafner2019dreamer}. Interactive models extend this idea to control-conditioned video~\citep{
yang2024learninginteractiverealworldsimulators,
bruce2024geniegenerativeinteractiveenvironments}.
Geometry-aware methods such as Spatia and WorldStereo further improve spatial consistency and camera control through 3Dguidance~\citep{zhao2025spatiavideogenerationupdatable,
zhang2026worldstereobridgingcameraguidedvideo}. 
Yet these systems emphasize camera-driven exploration, with limited modeling of physically grounded navigation actions. 
Our \method{} instead uses simulated robot actions to determine future motion and camera trajectories before visual prediction.

\subsection{3D Reconstruction toward Physical Scene Understanding}
Recent 3D reconstruction methods such as VGGT and MapAnything enable efficient scene reconstruction from images~\citep{wang2025vggt,keetha2026mapanything}.
Building on these geometric representations, IGGT and VGGT-Segmentor further incorporate instance-level and semantic understanding~\citep{li2026iggt,gao2026vggtsegmentor}. However,  physical scene properties are still  less explored. 
Recent VLM-based methods such as SiPhy and PhysGS begin to address this direction by estimating physical attributes from visual cues~\citep{le2026siphy,chopra2026physgs}. 
These approaches often rely on multi-stage inference, making efficient scene-wide physical prediction challenging.
\method{} targets this gap by extending feed-forward 3D reconstruction with dense physical priors.

\subsection{Robot Navigation Data}

Navigation datasets cover diverse domains. NuScenes targets urban driving~\citep{caesar2020nuscenes}, while RELLIS-3D and GOOSE focus on unstructured outdoor environments~\citep{jiang2022rellis3ddatasetdatabenchmarks,
mortimer2024goose}. 
ScanNet provides indoor geometry~\citep{dai2017scannet},
and SCAND records navigation behavior~\citep{9799755}. However,
most datasets remain domain-specific and capture limited physical variation
across terrain surfaces. 
Terrain perception estimates traversability and
physical properties~\citep{frey23fast,miki2022perceptive,
chen2024identifyingterrainphysicalparameters}, whereas real-to-sim methods
infer simulator parameters from observed responses~\citep{
chebotar2019closingsimtorealloopadapting,
ramos2019bayessimadaptivedomainrandomization}.
Our work connects visual understanding with executable physics for future prediction.
\section{Method}
\label{sec:method}

\subsection{Physical--Visual Prediction Framework}
\method{} predicts robot motion and future observations under terrain-dependent contact dynamics.
\emph{PhyMap} reconstructs geometry and dense physical priors from causal views.
Calibrated using real robot action-response data, these priors define executable terrain parameters.
Genesis~\citep{genesis2024} then simulates robot motion from this terrain, the current state, and commands.
The resulting camera trajectories and rendered controls guide video generation.
Future viewpoints thus arise from robot--terrain interaction rather than prescribed trajectories.

\begin{figure}[!htbp]
\centering
\includegraphics[width=\linewidth]{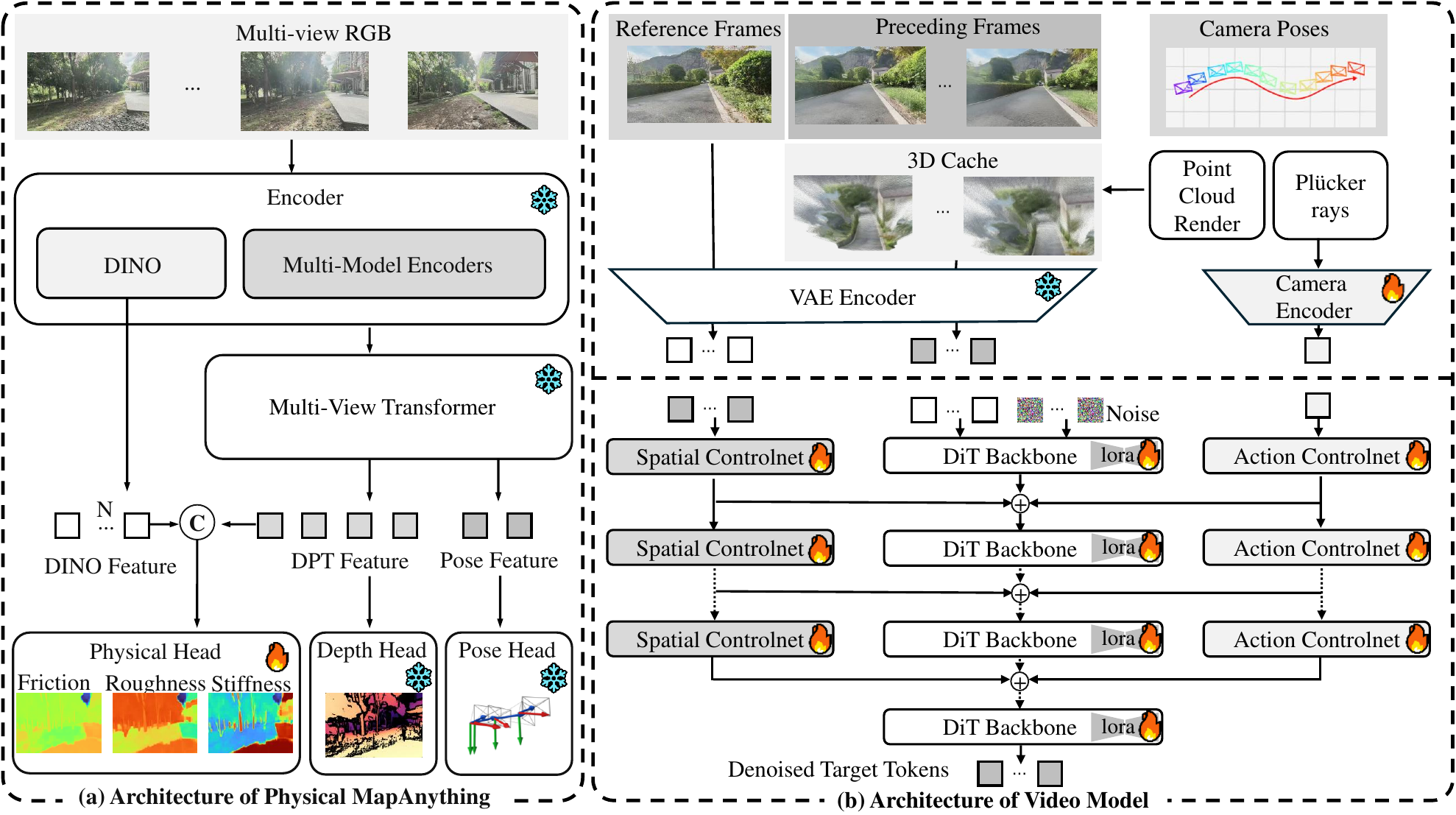}
\caption{Architecture of \method{}.
(a) \emph{PhyMap} reconstructs geometry and physical priors.
(b) Spatial and action ControlNets guide video generation using camera poses and rendered geometry.}
\label{fig:model-architecture}
\end{figure}

To formalize this physical--visual prediction, let $s_t$ denote the robot state, comprising the floating-base pose, velocity, and joint states.
Then $a_t$ denotes the recorded command and $I_t$ the synchronized robot-camera RGB frame.
At chunk boundary $\tau_k$, history $\mathcal H_k$ contains the initial real frame $\bar I_0=I_0$ and generated frames $\bar I_j=\hat I_j$ for $0<j\leq\tau_k$.
$\mathcal H_k$ also includes the current state and next $H$ commands, excluding future real frames.
The joint prediction is factorized as:
\begin{align}
&p\!\left(s_{\tau_k+1:\tau_k+H},I_{\tau_k+1:\tau_k+H}
\mid \mathcal H_k\right) \nonumber\\
&\quad =
p_{\Gamma}\!\left(s_{\tau_k+1:\tau_k+H}
\mid s_{\tau_k},a_{\tau_k:\tau_k+H-1},\Theta_k\right)
p_{\psi}\!\left(I_{\tau_k+1:\tau_k+H}
\mid \bar I_{0:\tau_k},\mathcal T_k,\mathcal C_k,\mathcal R_k\right),
\label{eq:factorization}
\end{align}
where $\Theta_k$ denotes the executable terrain and $\mathcal T_k$ the simulated camera trajectory.
The rendered 3D control and strictly past reference views are denoted by $\mathcal C_k$ and $\mathcal R_k$, respectively.
Genesis implements the physical factor deterministically, and the video model generates the corresponding future observations.
Figure~\ref{fig:model-architecture} summarizes the architecture.
Appendix~B provides calibration details.

\subsection{PhyMap: Dataset and Physical-Prior Learning}%Visual Physical World Reconstruction
\label{sec:phymap-method}
% PhyMap constructs a geometry-aligned physical representation with a frozen
% MapAnything backbone~\citep{keetha2026mapanything}:
% \begin{equation}
% \mathcal W_k^{\mathrm{phy}}
% =\mathcal P_{\omega}(\mathcal V_k)
% =\bigl\{(x_i,c_i,q_i,z_i)\bigr\}_{i=1}^{N_k},
% \qquad
% z_i=(z_i^f,z_i^r,z_i^s)\in[0,1]^3,
% \label{eq:phymap-representation}
% \end{equation}
% where $x_i\in\mathbb R^3$, $c_i$, and $q_i$ denote position, RGB, and geometric confidence, respectively.
% The three dimensionless channels represent friction,
% roughness, and stiffness priors, which are later mapped to simulator parameters.
% The frozen backbone provides multi-scale DPT~\citep{ranftl2021dpt} and DINOv2~\citep{oquab2024dinov2learningrobustvisual} features, along with depth,
% normals, and confidence. 
% A trainable physical decoder predicts $z_v(u)$ using a masked smooth-$L_1$ loss with auxiliary terrain classification.
% During training, PhyMap follows a view-count curriculum that progressively increases the number of causal input views.  
% The predicted physical attributes are lifted into 3D and aligned with the reconstructed point cloud, forming a geometry-aligned physical field. 
% To exclude non-support surfaces, only ground connected to the robot spawn region is retained, preventing walls from being interpreted as drivable terrain.
% \subsection{Visual Physical World Reconstruction}

\emph{PhyMap} constructs a geometry-aligned physical representation with a frozen
MapAnything backbone~\citep{keetha2026mapanything}:
\begin{equation}
\mathcal W_k^{\mathrm{phy}}
=\mathcal P_{\omega}(\mathcal V_k)
=\bigl\{(x_i,c_i,q_i,z_i)\bigr\}_{i=1}^{N_k},
\qquad
z_i=(z_i^f,z_i^r,z_i^s)\in[0,1]^3,
\label{eq:phymap-representation}
\end{equation}
where $x_i\in\mathbb R^3$, $c_i$, and $q_i$ denote position, RGB, and
geometric confidence, respectively.
The three dimensionless channels represent friction, roughness, and
stiffness priors, which are later mapped to simulator parameters.

To support \emph{PhyMap} training, we construct \emph{Physcene--Image}, a large-scale image dataset with dense physical-prior supervision, combining our real-world quadruped data, public indoor and off-road datasets, and web videos to cover diverse scenes. 
CLIP~\citep{radford2021learning} guides scene-based video filtering,
followed by frame extraction and deduplication.
We then use ActiveSAM~\citep{tien2026activesam} to segment all images
into a unified set of 32 semantic classes covering terrain, obstacles,
and traffic participants.
The complete dataset contains 2.58M unique RGB-mask pairs:
1.5M on-robot, 690K indoor, 44.6K off-road, and 350K web samples.
For physical supervision, we construct a class-to-property lookup
table from LLM priors.
The table converts segmentation masks into dense targets for three normalized physical properties.
Appendix~C provides the detailed source breakdown.

The frozen backbone provides multi-scale DPT~\citep{ranftl2021dpt}
and DINOv2~\citep{oquab2024dinov2learningrobustvisual} features,
along with depth, normals and confidence.
A trainable physical decoder predicts $z_v(u)$ at pixel $u$ in view $v$.
Supervision combines a masked smooth-$L_1$ loss on the physical targets
with auxiliary terrain classification.
Training follows a view-count curriculum that progressively increases
the number of causal input views.
The predicted priors are lifted into 3D to form a physical field aligned
with the reconstructed point cloud.

\subsection{Behavior-Grounded Executable Physics}
\label{sec:behavior-grounding-method}
The normalized physical field is mapped pointwise by $g_{\phi}$ to executable terrain parameters $\theta(x)=(\mu(x),\rho(x),\tau_c(x))$. Here $\mu$ denotes the ground friction coefficient, $\rho$ an executable roughness/contact-scale proxy, and $\tau_c$ the contact constraint time constant. 
Larger visual stiffness is mapped monotonically to a smaller $\tau_c$.
This parameterization builds on terrain physical-parameter perception~\citep{chen2024identifyingterrainphysicalparameters} and is calibrated from observed robot responses. 
The mapping uses monotone piecewise interpolation:
\begin{equation}
\mu=\operatorname{PL}(z^f;A_f),\qquad
\rho=\operatorname{PL}(z^r;A_r),\qquad
\tau_c=\exp\!\left(\operatorname{PL}
(z^s;\{(a,\log b):(a,b)\in A_s\})\right),
\label{eq:monotone-mapping}
\end{equation}
followed by clipping to the registered Genesis ranges.
In the held-out executable-prior experiment, $\rho$ controls the contact-impedance transition width, with the point-cloud collision height held fixed.

% Behavior grounding calibrates $g_{\phi}$ using synchronized real action-response snippets.
% For snippet $i$, $D_i$ measures the discrepancy between Genesis replay and normalized real trajectory, velocity, orientation, and impact responses:
% \begin{equation}
% \phi^*=\arg\min_{\phi}\frac{1}{|\mathcal D_{\mathrm{cal}}|}
% \sum_{i\in\mathcal D_{\mathrm{cal}}}D_i(g_{\phi}(z_i)).
% \label{eq:behavior-calibration}
% \end{equation}
% The resulting mapping combines a scene-level behavioral optimum with monotone endpoint priors, grounding the predicted physical channels in robot response.
% The robust discrepancy and identifiability scope are described in Appendix~B.

Behavior grounding identifies scene-level effective contact parameters
using synchronized real action-response snippets.
For snippet $i$, $D_i$ measures the normalized discrepancy between
Genesis replay and real trajectory, velocity, orientation, and impact
responses:
\begin{equation}
\theta^*=\arg\min_{\theta}\frac{1}{|\mathcal D_{\mathrm{cal}}|}
\sum_{i\in\mathcal D_{\mathrm{cal}}}D_i(\theta).
\label{eq:behavior-calibration}
\end{equation}
Here, $\mathcal D_{\mathrm{cal}}$ contains snippets from the calibration
scene. 
The resulting mapping combines this behavioral optimum with monotone endpoint priors, grounding the predicted physical channels in robot response.
The robust discrepancy and identifiability scope are described in Appendix~B.

\subsection{Video Generation Model Training}
Built on a Wan backbone~\citep{wan2025}, the generator integrates spatial and action control through two complementary branches. 
The spatial branch adopts coarse 3D conditioning~\citep{li2025wonderplaydynamic3dscene}: RGB and confidence controls are rendered from the causal point map, encoded by the frozen Wan VAE, and injected through
VACE~\citep{jiang2025vaceallinonevideocreation}. 
Spatial references are selected from $r<\tau_k$ using visible-point overlap, as in Spatia~\citep{zhao2025spatiavideogenerationupdatable}.
% In parallel, a dedicated action ControlNet directly conditions generation on robot commands.
In parallel, a dedicated action ControlNet conditions generation on camera poses derived from robot motion.
The dual-ControlNet design is introduced to maintain reliable action following under complex off-road terrain while preserving spatial consistency.
Detailed ablations are provided in Appendix~E.

Training proceeds in two stages.
The first stage optimizes the spatial VACE branch, input projection, and LoRA adapters in the attention and feed-forward projections~\citep{hu2021loralowrankadaptationlarge}.
The second stage introduces the action ControlNet and jointly optimizes it with the LoRA adapters, while preserving spatial control. 
Conditioned history and reference latents receive timestep zero, while the loss is applied only to future latent tokens. 
Appendix~B provides further training and conditioning details.
\subsection{Causal Autoregressive Rollout}
At chunk $k$, the complete executable update is:
\begin{align}
\mathcal W_k^{\mathrm{phy}}
&=\mathcal P_{\omega}(\{\bar I_j:j\in\mathcal J_k\}),
&&\mathcal J_k\subseteq\{0,\ldots,\tau_k\},\nonumber\\
\Theta_k&=g_{\phi}(\mathcal W_k^{\mathrm{phy}}),
&&s_{\tau_k+1:\tau_k+H}=F_{\Gamma}
(s_{\tau_k},a_{\tau_k:\tau_k+H-1};\Theta_k),\nonumber\\
\mathcal T_k&=\mathcal K(s_{\tau_k:\tau_k+H},T_{BC}),
&&\hat I_{\tau_k+1:\tau_k+H}
\sim p_{\psi}(\cdot\mid\bar I_{0:\tau_k},\mathcal T_k,
\mathcal C_k,\mathcal R_k).
\label{eq:causal-recurrence}
\end{align}
The $\mathcal J_k$ indexes only the initial frame and previously generated views, excluding future real observations.
Genesis resumes from its checkpoint, preserving state continuity across chunks.
Each new video chunk extends the fixed prefix and supplies views for the next PhyMap update.
Appendix~B details bootstrap and keyframe selection.

Through this causal autoregressive rollout, \method{} predicts terrain-dependent action consequences by iterating PhyMap reconstruction, Genesis simulation, and video generation.
The resulting videos provide future context for VLMs to compare candidate actions and inform decisions.

\section{Experiments}
\label{sec:experiments}
\subsection{Experimental Setup}
% Experiments assess executable physical priors and action-conditioned motion and future-observation prediction.
% To assess the utility of \method{}, we conduct VLM action selection experiments.
% Appendices cover ablations, physical-prior remapping, causal long-horizon analyses, and independent-scene Genesis-in-the-loop motion prediction.
% Appendix~\ref{app:phymap} details VLM segmentation--lookup priors.

% Evaluation data include held-out scenes from our robot collection and public datasets.
% Real-robot data are collected using DEEPRobotics M20 and Lite3  robots.
% Scene-level partitions are disjoint, and the behavior-calibrated PhyMap mapping remains fixed at test time.
% Comparisons are paired by scene.
% Physics ablations hold geometry, initial state, actions, controller, simulator, video model, seed, and duration fixed.
% Future real RGB frames are excluded from causal inference, and only evaluation outputs are resampled.
% Confidence intervals, where reported, are computed by bootstrapping scenes.
% Appendix~A provides further evaluation details.

Experiments evaluate executable physical priors, action conditioned motion, and future observation prediction.
We further examine their utility for VLM action selection.
The appendices report additional ablations, physical prior remapping, long horizon causal rollouts, and Genesis in the loop motion prediction.
Appendix~\ref{app:phymap} describes the VLM segmentation and lookup baseline.

% Evaluation covers held-out scenes from our dedicated DEEPRobotics M20/Lite3
% dataset, which synchronizes RGB observations, robot states, and control commands.
% Scene-level partitions are disjoint, and the behavior-calibrated PhyMap mapping remains fixed at test time.
% Tables~\ref{tab:action_to_observation}, \ref{tab:reference-fidelity},\ref{tab:cross-dataset}, and \ref{tab:spatial-prior-ablation}
% all use held-out on-robot scenes.
% Physics ablations keep geometry, initial states, actions, controllers, simulators, video models, seeds, and durations unchanged.
% Causal inference never accesses future real RGB frames.
% Resampling applies only to evaluation outputs.
% We compute reported confidence intervals with scene bootstrap.
% Appendix~A provides further evaluation details.

Evaluation uses held-out scenes from \emph{Physcene--Robot}, the robot-interaction subset of \emph{Physcene}.
\emph{Physcene--Robot} contains over 100K synchronized M20/Lite3 clips with RGB observations, robot states, and control commands.
Scene-level partitions are disjoint, and the behavior-calibrated \emph{PhyMap} mapping remains fixed at test time.
Tables~\ref{tab:action_to_observation}, \ref{tab:reference-fidelity},
\ref{tab:human-preference}, and \ref{tab:spatial-prior-ablation}
all use held-out \emph{Physcene--Robot} scenes.
% Physics ablations keep geometry, initial states, actions, controllers, simulators, video models, seeds, and durations unchanged.
% Future real RGB frames are never used by non-oracle causal inference.
% Resampling applies only to evaluation outputs.
% Reported confidence intervals are computed by bootstrapping scenes.
% Appendix~A provides further evaluation details.
The executable-prior ablation in Table~\ref{tab:spatial-prior-ablation} varies only terrain contact parameters, keeping all other settings fixed.
Non-oracle rollouts exclude future real RGB frames from their inputs.
Resampling applies only to evaluation outputs.
Reported confidence intervals are computed by bootstrapping scenes.
Appendix~A provides further evaluation details.

\subsection{Comparison with Action-Conditioned World Models}
\label{sec:exp-action-observation}

\begin{figure}[!htbp]
\centering
\includegraphics[width=\linewidth]{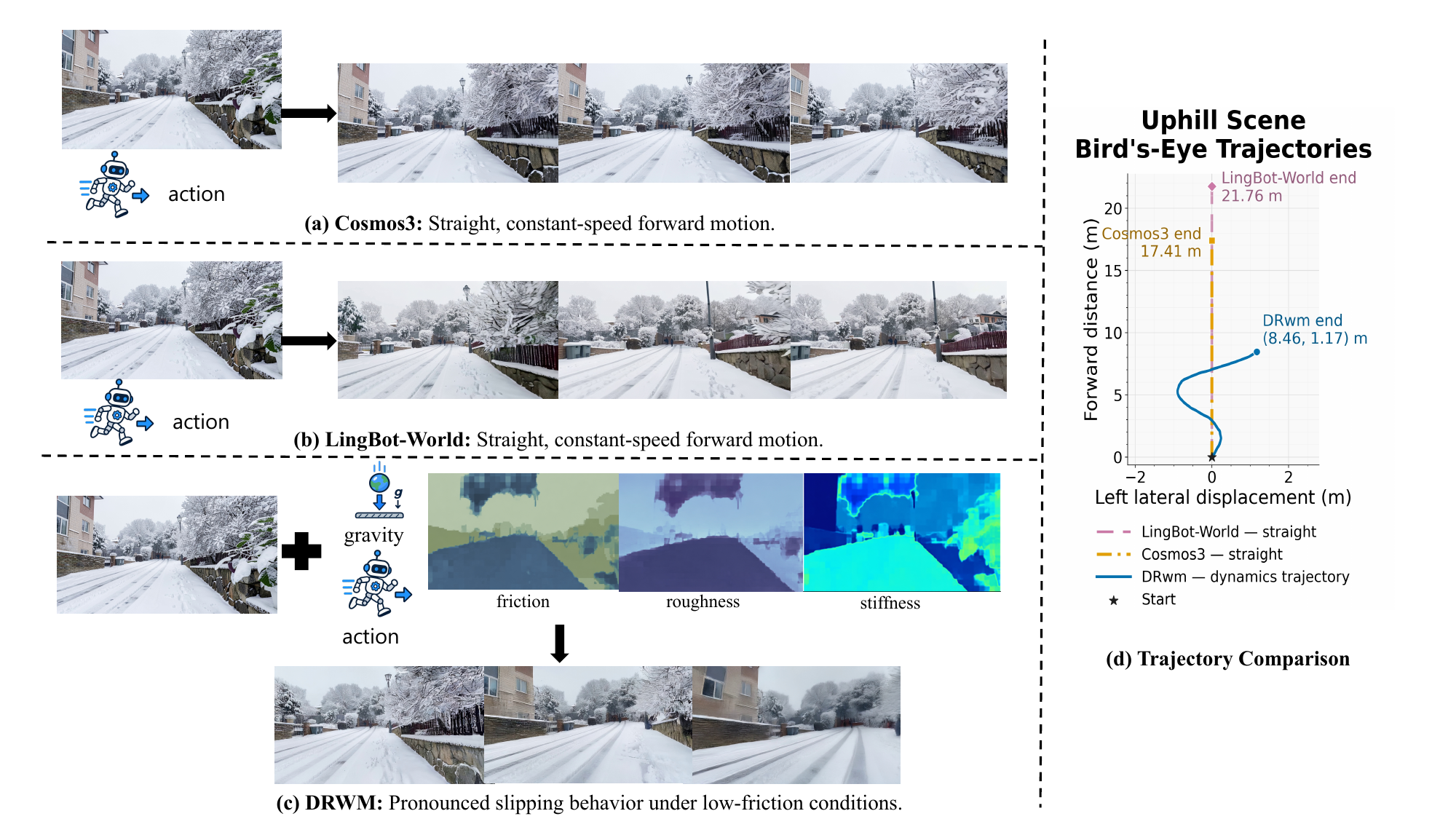}
\caption{Qualitative comparison on an uphill snowy scene. Cosmos~3 and LingBot-World follow straight, constant-speed motion conditions, whereas \method{} predicts a dynamics-mediated trajectory from the action and inferred low-friction terrain properties. The right panel compares the resulting bird's-eye trajectories.}
\label{fig:action-observation-main}
\end{figure}

To evaluate action-conditioned world models, we compare \method{} with representative baselines.
The comparison includes image-to-video, camera-controlled, and action-conditioned methods.
We compare CogVideoX1.5~\citep{yang2025cogvideoxtexttovideodiffusionmodels},
Wan2.2-TI2V~\citep{wan2025},
WorldStereo~\citep{zhang2026worldstereobridgingcameraguidedvideo},
Matrix-Game~3.0~\citep{wang2026matrixgame30realtimestreaming},
the Cosmos~3 AV configuration~\citep{aditi2026cosmos3},
EgoWM~\citep{bagchi2026walkpaintingsegocentricworld}, LingBot-World-Base~\citep{
robbyantteam2026advancingopensourceworldmodels}, and NWM~\citep{
bar2025navigationworldmodels}.
All methods share the same initial frame and are evaluated against the same
recorded future.
Controllable baselines receive recorded actions or measured camera poses;
image-to-video baselines receive a matched prompt.
\method{} receives the initial robot state and commands and predicts robot/camera motion through Genesis (Figure~\ref{fig:action-observation-main}).

% Table~\ref{tab:action_to_observation} uses fixed evaluators for imaging,
% aesthetics, motion, temporal consistency, action adherence, and physical
% realism, following multidimensional video evaluation~\citep{
% li2025wonderplaydynamic3dscene,huang2023vbenchcomprehensivebenchmarksuite}.
% The first four measure perceptual quality and temporal coherence; action
% adherence measures agreement with the recorded future, and physical realism assesses robot-terrain interaction. 
% Prompts, aggregation, and failure handling are fixed.

% \input{tables/tab_action_to_observation_main}

% \method{} leads all six criteria in Table~\ref{tab:action_to_observation}.
% Relative to Matrix-Game~3.0, the strongest baseline for action adherence and
% physical realism,
% \method{} raises action adherence from 0.913 to 0.949 and physical realism
% from 0.870 to 0.913.

Table~\ref{tab:action_to_observation} evaluates generated videos on 50  \emph{Physcene--Robot} scenes using GPT-5.5~\citep{openai2026gpt55} as the evaluator.
Following multidimensional video evaluation~\citep{
li2025wonderplaydynamic3dscene,huang2023vbenchcomprehensivebenchmarksuite}, we consider imaging quality, aesthetics, motion quality, temporal consistency, action adherence, and physical realism.
The first four assess visual quality and temporal coherence, while the latter two capture agreement with the commanded motion and robot--terrain interaction.
All methods share the same prompts and evaluation protocol.
\begin{table}[!htbp]
\centering
\renewcommand{\arraystretch}{1.1}
\caption{VLM-based automatic evaluation of generated video quality on on-robot scenes.}
\label{tab:action_to_observation}
\setlength{\tabcolsep}{3pt}
\begin{tabular}{@{}lcccccc@{}}
\toprule
Method & Imaging$\uparrow$ & Aesthetic$\uparrow$ & Motion$\uparrow$ & Temp. Cons.$\uparrow$ & Action Adh.$\uparrow$ & Phys. Realism$\uparrow$ \\
\midrule
CogVideoX1.5 & 0.736 & 0.742 & 0.826 & 0.781 & 0.839 & 0.793 \\
Wan2.2-TI2V & 0.779 & 0.774 & 0.859 & 0.846 & 0.884 & 0.842 \\
WorldStereo & 0.783 & 0.763 & 0.875 & 0.861 & 0.892 & 0.858 \\
Matrix-Game 3.0 & 0.789 & 0.782 & 0.886 & 0.864 & 0.913 & 0.870 \\
Cosmos~3 & 0.797 & 0.791 & 0.881 & 0.873 & 0.908 & 0.864 \\
EgoWM & 0.624 & 0.653 & 0.798 & 0.669 & 0.801 & 0.706 \\
LingBot-World & 0.673 & 0.666 & 0.419 & 0.714 & 0.334 & 0.328 \\
NWM & 0.243 & 0.287 & 0.094 & 0.118 & 0.041 & 0.069 \\
\textbf{\method{} (Ours)} & \textbf{0.814} & \textbf{0.812} & \textbf{0.924} & \textbf{0.896} & \textbf{0.949} & \textbf{0.913} \\
\bottomrule
\end{tabular}
\end{table}

\method{} performs best on all six criteria, improving action adherence from 0.913 to 0.949 and physical realism from 0.870 to 0.913 over Matrix-Game~3.0.

Table~\ref{tab:reference-fidelity} complements these primarily reference-free
scores with reference comparisons: PSNR (pixel fidelity),
LPIPS~\citep{zhang2018unreasonableeffectivenessdeepfeatures} and
DINO similarity~\citep{caron2021emergingpropertiesselfsupervisedvision}
(perceptual/structural agreement), RAFT flow error~\citep{
teed2020raftrecurrentallpairsfield}, and FVD-16~\citep{
unterthiner2019accurategenerativemodelsvideo} (short-video distributions).

A frozen estimator recovers video camera poses for RotErr and ATE against
recorded trajectories. Evaluation uses a common resolution and 17 aligned
positions, excluding the initial frame. WorldStereo uses its sparse-trajectory
timestamps; other baselines use the common four-second window.
\begin{table}[!htbp]
\centering
\renewcommand{\arraystretch}{1.1}
% \caption{Reference-aligned fidelity. RotErr and ATE compare camera poses recovered from generated videos with the recorded trajectories; frame metrics use corresponding real frames, and FVD-16 measures short-video distributional fidelity.}
\caption{Reference aligned evaluation of camera motion and video fidelity.}
\label{tab:reference-fidelity}
\setlength{\tabcolsep}{2pt}
\begin{tabular*}{\linewidth}{@{\extracolsep{\fill}}lcc|ccccc@{}}
\toprule
Method & RotErr ($^\circ$)$\downarrow$ & ATE (m)$\downarrow$ & PSNR$\uparrow$ & LPIPS$\downarrow$ & DINO Sim.$\uparrow$ & Flow EPE$\downarrow$ & FVD-16$\downarrow$ \\
\midrule
CogVideoX1.5 & 6.357 & 20.497 & 13.68 & 0.4810 & 0.6082 & 21.546 & 738.37 \\
Wan2.2-TI2V & 7.717 & 11.860 & 13.53 & 0.4272 & 0.6990 & 18.819 & 683.42 \\
WorldStereo & 7.060 & 6.395 & 13.24 & 0.4843 & 0.6213 & 22.199 & 599.33 \\
Matrix-Game 3.0 & 4.923 & 1.704 & 14.27 & 0.4623 & 0.6902 & 9.053 & 1199.86 \\
Cosmos~3 & 20.805 & 3.653 & 15.09 & 0.3987 & 0.7441 & 11.992 & 801.40 \\
EgoWM & 7.469 & 1.073 & 15.24 & 0.4317 & 0.6377 & 8.592 & 824.32 \\
LingBot-World & 13.167 & 13.514 & 13.58 & 0.5020 & 0.6021 & 16.329 & 1148.22 \\
NWM & 67.217 & 14.852 & 12.20 & 0.6151 & 0.2079 & 88.469 & 1525.20 \\
\textbf{\method{} (Ours)} & \textbf{3.980} & \textbf{0.961} & \textbf{17.01} & \textbf{0.2818} & \textbf{0.7993} & \textbf{6.001} & \textbf{567.27} \\
\bottomrule
\end{tabular*}
\end{table}

In a blind study, 100 volunteers rated anonymized clips, given a common initial frame and target-motion reference. 
Mean 0--100 ratings for action/physical plausibility, motion fidelity, temporal
consistency, and visual quality appear in Table~\ref{tab:human-preference}.
\method{} leads by 9.3, 7.6, 8.8, and 5.5 points over the strongest baseline,
respectively.

\begin{table}[!htbp]
\centering
\renewcommand{\arraystretch}{1.1}
\caption{Blind human evaluation on M20/Lite3 clips. Cells are mean ratings on a 0--100 scale.}
\label{tab:human-preference}
\begin{tabular}{@{}lcccc@{}}
\toprule
Method & Act./Phys. Plaus.$\uparrow$ & Motion Fid.$\uparrow$ & Temp. Cons.$\uparrow$ & Visual Qual.$\uparrow$ \\
\midrule
CogVideoX1.5 & 74.6 & 76.8 & 75.9 & 77.2 \\
Wan2.2-TI2V & 83.8 & 81.5 & 81.2 & 85.3 \\
Cosmos~3 & 82.4 & 84.1 & 83.6 & 84.7 \\
LingBot-World & 51.8 & 56.3 & 60.4 & 66.8 \\
\textbf{\method{} (Ours)} & \textbf{93.1} & \textbf{91.7} & \textbf{92.4} & \textbf{90.8} \\
\bottomrule
\end{tabular}
\end{table}

\subsection{Held-Out Executable-Prior Ablation}
\label{sec:exp-executable}

We evaluate 34 clips from 22 held-out scenes of \emph{Physcene--Robot}
with frozen \emph{PhyMap} and $g_\phi$. 
All variants share collision geometry, initial state, action, controller, and simulator.
\emph{Geometry Only} uses default physical parameters.
\emph{Uniform PhyMap} broadcasts each calibrated contact-parameter field's valid-terrain mean, while \emph{Full PhyMap} retains spatial variation. 
Thus, \emph{Uniform PhyMap} removes spatial variation while preserving each parameter's mean. The metrics cover base trajectory, relative vertical motion, roll/pitch, and velocity. 
Vertical RMSE compares predicted and ground-truth vertical displacements in their respective initial base frames.
And velocities are computed using the same pose-differentiation procedure.

\begin{table}[!htbp]
\centering
\renewcommand{\arraystretch}{1.1}
% \caption{Executable-prior ablation exclusively on 34 held-out clips from 22 scenes of our dedicated M20/Lite3 dataset. Values are scene means. All variants share collision geometry; Uniform PhyMap broadcasts each field's valid-terrain mean, whereas Full PhyMap retains spatial variation.}
\caption{Executable physical prior ablation on on-robot scenes.}
\label{tab:spatial-prior-ablation}
\setlength{\tabcolsep}{5pt}
\begin{tabular}{@{}lcccc@{}}
\toprule
Method & Base ATE (m)$\downarrow$ & Vert. RMSE (m)$\downarrow$ & R/P RMSE ($^\circ$)$\downarrow$ & Vel. RMSE (m/s)$\downarrow$ \\
\midrule
Geometry Only & 0.1683 & 0.0830 & 8.0215 & 1.2238 \\
Uniform PhyMap & 0.1450 & 0.0500 & 4.8269 & 1.0850 \\
\textbf{Full PhyMap} & \textbf{0.1380} & \textbf{0.0378} & \textbf{3.3553} & \textbf{1.0246} \\
\bottomrule
\end{tabular}
\end{table}

Table~\ref{tab:spatial-prior-ablation} reports the held-out results. Compared with Geometry Only, Full \emph{PhyMap} reduces base ATE by 18.0\%, vertical-motion RMSE by 54.5\%, roll/pitch RMSE by 58.2\%, and velocity RMSE by 16.3\%. Compared with Uniform \emph{PhyMap}, retaining spatial variation reduces base ATE by 4.8\%, vertical-motion RMSE by 24.4\%, roll/pitch RMSE by 30.5\%, and velocity RMSE by 5.6\%, giving Full \emph{PhyMap} the best mean performance on all four metrics.

\subsection{VLM Action Selection with Closed-Loop Physical Rollouts}
\label{sec:exp-vlm-decision}

We evaluate VLM action selection on 90 challenging scenes with candidate-wise
closed-loop physical--visual rollouts.
Qwen3-VL-8B and Qwen3-VL-32B~\citep{bai2025qwen3vl} share the current image, goal, candidate
descriptions, and output specification. Their future context is either absent
(\emph{None}), generated by Wan (\emph{Generic WM}), or provided by
\method{} as time-stamped frames (Figure~\ref{fig:vlm-action-selection}).
For the selected branch, we measure goal progress (Prog.), path efficiency
(Eff.), world-up tilt RMS, mean tilt over the highest-tilt 10\% (Tilt T10),
logged contact-point slip (Slip), and arrival time. Bold/underlined values in
Table~\ref{tab:vlm-physical-rollout} mark best/second-best results across both sizes.

\begin{table}[!htbp]
\centering
\renewcommand{\arraystretch}{1.1}
\caption{VLM action selection on 90 challenging scenes with no future context, generic world-model rollouts, or DR-WM's closed-loop physical--visual rollouts.}
\label{tab:vlm-physical-rollout}
\setlength{\tabcolsep}{3pt}
\begin{tabular}{@{}clcccccc@{}}
\toprule
VLM & Rollout model & Prog.$\uparrow$ & Tilt RMS ($^\circ$)$\downarrow$ & Tilt T10 ($^\circ$)$\downarrow$ & Slip (m/s)$\downarrow$ & Arrival (s)$\downarrow$ & Eff.$\uparrow$ \\
\midrule
                        & None  & 0.7492 & 2.5760 & 4.8268 & 0.1580 & \textbf{7.3188} & 0.7912 \\
Qwen3-VL-8B             & Generic WM & 0.8442 & 3.1755 & 6.3250 & 0.1772 & 7.6076 & 0.8097 \\
                        & DR-WM & \underline{0.8754} & \underline{2.2590} & \underline{4.1889} & \textbf{0.1317} & 7.5310 & 0.8980 \\
\midrule
                        & None  & 0.8586 & 2.5856 & 5.1715 & 0.1675 & 7.5415 & \underline{0.9284} \\
Qwen3-VL-32B            & Generic WM & 0.8465 & 2.6262 & 5.0278 & 0.1556 & 7.5785 & 0.9130 \\
                        & DR-WM & \textbf{0.8921} & \textbf{2.1871} & \textbf{4.0894} & \underline{0.1351} & \underline{7.3338} & \textbf{0.9699} \\
\bottomrule
\end{tabular}
\end{table}

DR-WM leads five metrics for 8B and all six for 32B. For 8B, progress rises
from 0.7492 to 0.8754 and efficiency from 0.7912 to 0.8980 over no future
context; unassisted 8B remains fastest (7.3188\,s). Generic WM improves 8B
progress but worsens tilt and slip.
With DR-WM assistance, the 8B model achieves better mean results than
the unassisted 32B model on five of the six metrics. Upper-tail tilt and
slip decrease by 19.0\% and 21.4\%; mean arrival time improves by only
0.0105\,s. Path efficiency is the exception (0.8980 vs.\ 0.9284).

\section{Conclusion}
\label{sec:conclusion}

We presented DR-WM, a behavior-grounded world model for robot--terrain interaction
that grounds video prediction in executable physics calibrated from real
action--response data. By using physical simulation to guide video generation
and generated observations to update terrain reconstruction, DR-WM supports
continued physical--visual rollout for comparing candidate actions. Its
spatially varying terrain physics improves held-out motion prediction over
geometry-only and uniform-physics alternatives, while DR-WM leads the reported
automatic, reference-aligned, and human evaluations. Notably, its predicted
future observations enable an 8B VLM to achieve better mean
results than an unassisted 32B model on five of six action-selection metrics.

Despite these results, reliable prediction depends on the accuracy of
reconstruction, contact modeling, and generated observations. Because generated
observations update the terrain reconstruction, prediction errors may propagate
across successive rollouts. Improving geometric consistency and contact modeling
over longer horizons is therefore an important direction for future work. More
broadly, our results highlight the value of grounding visual world models in
measured robot responses. Connecting commanded actions to terrain-dependent
motion and its visual consequences offers a path toward more informed action
selection before real-world execution.

% The appendix is included below so this file directly builds DRWM.pdf.

% DEEP Robotics template natbib/BibTeX bibliography setup.
% Include works cited only in the separately compiled supplementary material.
\nocite{allevato2020tunenetoneshotresidualtuning,chopra2026physgs,
  kuhn1955hungarian,le2026siphy,qwen38,shoemake1985,shuai2025pugs,
  sobanbabu2025samplingbasedidentificationactiveexploration,
  yu2017preparingunknownlearninguniversal}
\par
\bibliography{references}
\bibliographystyle{plainnat}

\par\bigskip
% \setcounter{page}{1}
% \setcounter{figure}{0}
% \setcounter{table}{0}
% \setcounter{equation}{0}

% Preserve the prefixed cross-reference names used by the main paper.
\let\drwmoriginallabel\label
\renewcommand{\label}[1]{%
  \drwmoriginallabel{#1}%
  \ifstrequal{#1}{app:phymap}{\drwmoriginallabel{supp-app:phymap}}{}%
  \ifstrequal{#1}{tab:strict-genesis-rollout}{%
    \drwmoriginallabel{supp-tab:strict-genesis-rollout}}{}%
}

\section*{APPENDIX}

\appendix

\begin{figure}[!htbp]
\centering
\includegraphics[
  page=7,
  width=\linewidth,
  trim=124bp 6bp 124bp 0bp,
  clip
]{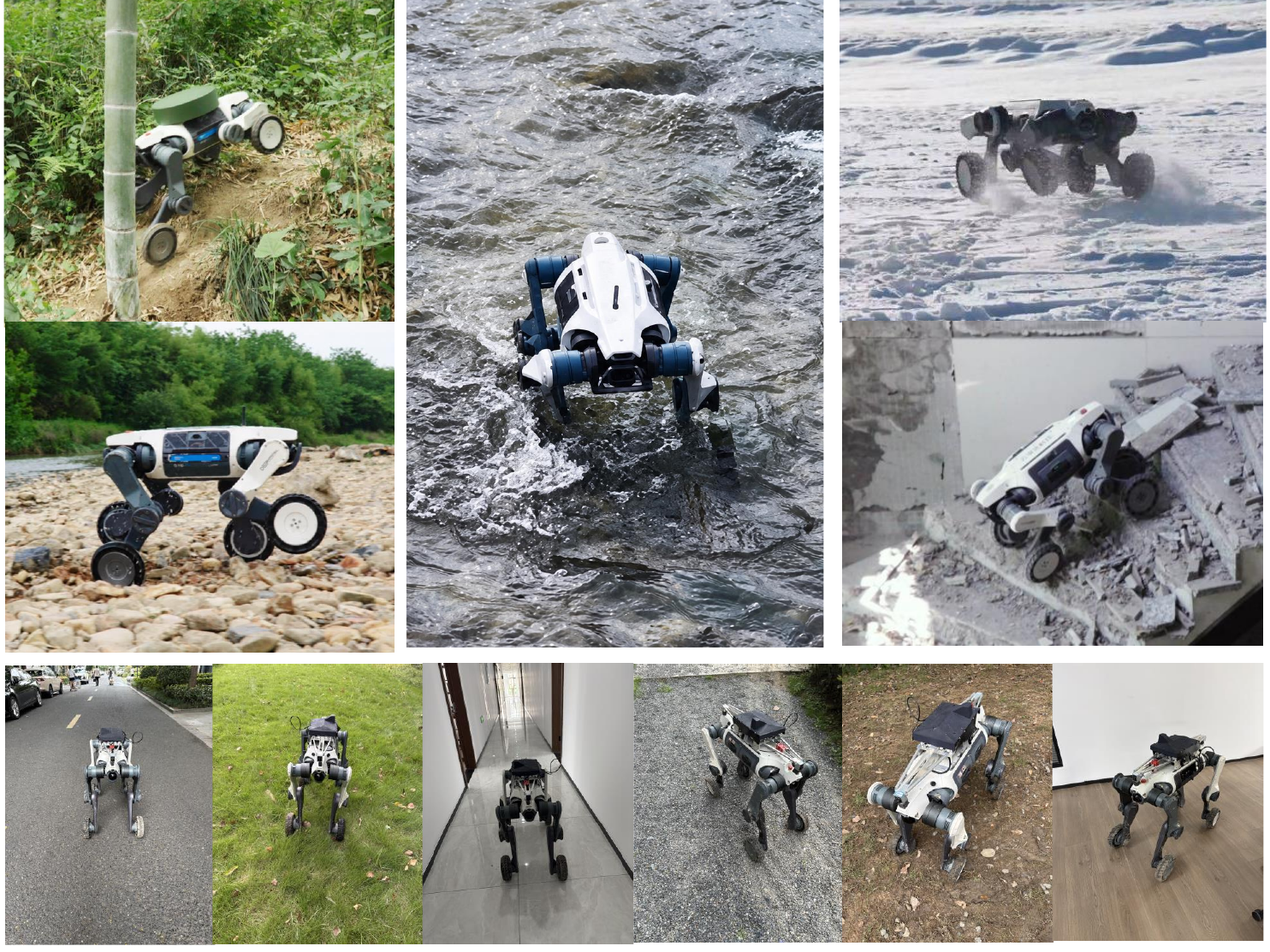}
\caption{Illustration of robot data collection and testing.}
\label{fig:dedicated-dataset}
\end{figure}

\section{Datasets, Physical Labels, and Implementation Details}
\label{app:datasets_impl}

\subsection{Physcene--Robot Dataset}

% We collected a dedicated dataset with real DEEP Robotics M20 and Lite3 robots
% to evaluate the paper's central action-to-executable-physics-to-observation
% pipeline under real embodied interactions. Each recorded sequence associates
% timestamped robot-mounted RGB camera images and LiDAR scans with the robot's
% motion/state response and the control commands issued during execution. The
% camera stream supplies the visual observations, while LiDAR provides metric
% geometry for supervision, registration, and evaluation. The synchronized
% action-response record contains the executed robot motion and the corresponding
% control inputs; in particular, the raw six-field \texttt{/STEER} samples are
% converted to the forward, lateral, and yaw commands consumed by the robot-policy
% adapter described in Appendix~\ref{app:method_details}.
\emph{Physcene--Robot} is a dedicated real-robot interaction dataset collected with DEEPRobotics M20 and Lite3 robots.
It is designed to support behavior grounding and evaluation of the
"Action--Executable Physics--Observation" pipeline.
Each sequence synchronizes timestamped robot-mounted RGB images and LiDAR scans with robot states, motion responses, and control commands.
RGB provides visual observations, while LiDAR supplies metric geometry for supervision, registration, and evaluation.
The recorded action-response data capture both executed motion and control inputs.
Raw six-field \texttt{/STEER} commands are converted to forward, lateral, and yaw commands through the robot-policy adapter described in Appendix~\ref{app:method_details}.

Camera--LiDAR and camera-base calibrations place observations, robot states,
and reconstructed terrain in a common coordinate system. We divide continuous
recordings into scene-level clips and keep temporally adjacent clips in the same
partition. 
Future RGB and LiDAR observations serve as training supervision or evaluation
ground truth and are excluded from causal inference.
Main-paper evaluations in Tables~\ref{tab:action_to_observation}--~\ref{tab:spatial-prior-ablation} use held-out scenes from this
dedicated \emph{Physcene--Robot} dataset. Figure~\ref{fig:dedicated-dataset} illustrates the range of robots,
terrains, and indoor/outdoor environments represented in our collection. The
outdoor environments cover snow, gravel, asphalt, grass, mud, and stone-paved
surfaces, while the designed indoor scenes include tile, wood flooring, and
high-pile carpet. The dataset contains more than 100,000 clips, with a phased
public release planned. It complements robot navigation
datasets by combining diverse indoor and outdoor terrains with synchronized
perception, control, and robot-response measurements. Each clip includes surround-view
video, dual-LiDAR geometry, inertial and joint states, odometry, motion state, and sensor calibration.

\paragraph{Representative raw-data inventory.}
We characterize the recorded sensor streams using one unprocessed 30~s MCAP
segment from the collection. As summarized in Table~\ref{tab:mcap-demo-inventory},
the representative \emph{Physcene--Robot} 30~s MCAP clip contains 20,756 messages across 12 active streams, stored in 177 indexed chunks. The embedded calibration attachment is marked valid and places the four fisheye cameras and two LiDARs in the common body-FLU reference frame.

\begin{table}[!htbp]
\centering
\caption{Representative sensor and control inventory from \emph{Physcene--Robot}.
The 30~s MCAP recording stores calibration as an attachment.}
\label{tab:mcap-demo-inventory}
\resizebox{\linewidth}{!}{%
\begin{tabular}{@{}llcc l@{}}
\toprule
Signal group & ROS~2 type / encoding & Rate (Hz) & Messages & Recorded content \\
\midrule
Fisheye cameras & \texttt{CompressedVideo}/H.264 & $\sim$10 & $4\!\times\!300$ & 1920$\times$1536 surround-view video \\
LiDAR & \texttt{PointCloud2}/CDR & $\sim$10 & $2\!\times\!300$ & $x,y,z$, intensity, ring, point time \\
Inertial sensing & \texttt{Imu}/CDR & $\sim$200 & 5,994 & Orientation, angular velocity, acceleration \\
Joint feedback & \texttt{JointsData}/CDR & $\sim$200 & 6,001 & 16-joint position, torque, velocity, temperature, status \\
LIO odometry & \texttt{Odometry}/CDR & $\sim$200 & 6,000 & 6-DoF pose, velocity, covariance \\
Motion state & \texttt{MotionInfo}/CDR & $\sim$20 & 600 & Velocity, yaw rate, height, state, gait, payload \\
Control input & \texttt{Steer}/CDR & $\sim$10 & 301 & $x,y,z$, roll, pitch, yaw command \\
Satellite navigation & \texttt{NavSat}/CDR & $\sim$2 & 60 & Position, speed, altitude, fix, satellites, DOP \\
Calibration & YAML attachment & --- & 1 & Camera intrinsics/extrinsics and LiDAR extrinsics \\
\bottomrule
\end{tabular}}
\end{table}

\subsection{Splits and Evaluation Controls}

We use scene-level train/validation/test partitions: adjacent clips remain in the
same partition, and behavior calibration, validation, and final test scenes are
disjoint. At test time, the PhyMap--Genesis mapping is frozen. All controlled
comparisons share the initial state, action sequence, controller, simulator,
duration, and sampling settings unless stated otherwise.

Primary comparisons are paired by scene. Where confidence intervals are
reported, bootstrap resampling uses scenes as the sampling unit. The
in-domain video-condition ablation in Appendix~\ref{app:video_ablation} is
reported separately from the held-out evaluations. The planned data and code
release includes split manifests, label definitions, sensor calibration,
robot/controller settings, Genesis parameters, video-model configurations, and
metric implementations.

\section{Detailed Method and Implementation}
\label{app:method_details}

This section expands the architecture, objectives, coordinate conventions, and
rollout rules summarized in Section~3 of the main paper.

\subsection{PhyMap Architecture and Executable-Terrain Construction}

For view $v$, the frozen MapAnything backbone extracts four DPT feature
levels~\citep{ranftl2021dpt} $\{F_v^{\ell}\}_{\ell=1}^4$ and a DINOv2
feature~\citep{oquab2024dinov2learningrobustvisual} $E_v$. Its geometric
prediction forms a five-channel per-pixel descriptor $d_v(u)$ comprising
log-normalized depth, the surface normal obtained from neighboring 3D-point
differences, and normalized log confidence. A trainable decoder $h_{\omega}$
performs top-down multi-scale fusion, local geometry cross-attention,
DINO-DPT fusion, and windowed self-attention; a sigmoid maps its per-pixel
output to $z_v(u)\in[0,1]^3$. Only $\omega$ is trained.

If $y_v(u)$ is a Terrain32 label, $m(y)\in[0,1]^3$ is its physical target from
the fixed terrain table, and $w_v(u)$ combines label, external, and
MapAnything-validity masks, the implemented objective is
\begin{align}
\mathcal L_{\mathrm{Phy}}
&=\frac{1}{\sum_{v,u}w_v(u)}
\sum_{v,u}w_v(u)
\left[
\sum_{d\in\{f,r,s\}}\rho_{\mathrm{SL1}}
\!\left(z_v^d(u)-m^d(y_v(u))\right)
\right.\nonumber\\[-0.2em]
&\hspace{8.2em}\left.
+\lambda_{\mathrm{terr}}\,
\operatorname{CE}(\ell_v(u),y_v(u))
\right],
\qquad \lambda_{\mathrm{terr}}=0.03,
\label{eq:app-phymap-loss}
\end{align}
where $\ell_v$ is an auxiliary terrain logit used only for supervision. The
deployed v7 checkpoint uses a 384-channel decoder, eight attention heads,
$8\times8$ windows, and three self-attention blocks.

To construct the executable 2.5D terrain, OpenCV camera coordinates are mapped
to Genesis coordinates as
$(x_{\Gamma},y_{\Gamma},z_{\Gamma})=(z_{\mathrm{cv}},-x_{\mathrm{cv}},-y_{\mathrm{cv}})$.
Within each terrain cell, we take a lower height quantile and the median of each
physical channel. We retain the four-connected component reachable from the
robot spawn region under a maximum adjacent-height change, fill unobserved
cells from the nearest retained ground cell, and smooth the resulting height
field. This prevents observed walls from being promoted to drivable support.

\subsection{Behavior Calibration Details}

Let $F_{\Gamma}$ denote a Genesis replay using the same initial state, recorded
command, controller, and duration as snippet $i$. For response measures $d_j$
with fixed unit normalizers $\kappa_j$ and weights $\lambda_j$, the
calibration discrepancy is
\begin{equation}
D_i(\theta)=\sum_j\lambda_j
\min\!\left(
\frac{d_j(F_{\Gamma}(s_{i,0},a_i;\theta),s_i^{\mathrm{real}})}
{\kappa_j},4\right),
\qquad \sum_j\lambda_j=1.
\label{eq:app-behavior-discrepancy}
\end{equation}
The terms cover planar trajectory, forward velocity, yaw and yaw rate, vertical
vibration/impact, and roll-pitch response. 
The scene-level optimum $\theta^*$ is obtained from Eq.~(\ref{eq:behavior-calibration}).
The mapping uses $g_\phi(z^*)=\theta^*$ as an anchor, where $z^*$ is the representative physical prior of the calibration scene.
Monotone endpoint priors complete the interpolant.
The normalizers, weights, and mapping remain fixed during test evaluation.
% The mapping uses a scene-level behavior optimum $\theta^*$ as the anchor $g_{\phi}(z^*)=\theta^*$, with monotone endpoint priors completing the interpolant. The normalizers, weights, and mapping remain fixed during test evaluation.
This formulation follows response-based real-to-sim calibration and system
identification~\citep{chebotar2019closingsimtorealloopadapting,
ramos2019bayessimadaptivedomainrandomization,
allevato2020tunenetoneshotresidualtuning,
yu2017preparingunknownlearninguniversal,
sobanbabu2025samplingbasedidentificationactiveexploration}. The resulting
parameters encode behavior-calibrated simulator responses; a behavior optimum
need not uniquely identify the underlying material constants.

\subsection{Robot, Camera, and Video-Model Details}

The six-field recorded \texttt{/STEER} sample is reduced to forward, lateral,
and yaw commands before the affine deployment adapter. The deployed policy
produces desired leg positions and wheel velocities. Genesis uses a
$0.005$-second step. Camera positions are linearly interpolated and rotations
are interpolated by quaternion SLERP~\citep{shoemake1985} to the video
timestamps.

For the visual branch, the frozen Wan VAE encodes rendered RGB control
$C_j^{\mathrm{rgb}}$ and geometric confidence $C_j^q$. Each encoding has 48
channels, yielding the 96-channel VACE context. For history-mode chunks, let
$V(r)$ be the set of frontmost map points visible in candidate reference $r$
and $V(j)$ the corresponding set for target $j$. Hungarian one-to-one
assignment~\citep{kuhn1955hungarian} maximizes
$\sum_j |V(\pi(j))\cap V(j)|$ subject to $\pi(j)<\tau_k$. We retain at most
seven references; the bootstrap chunk contains none.

Let $N_{\mathrm{hist}}\in\{1,9\}$ denote the number of RGB history frames ending at $\tau_k$. Define the clean training latent as $Z=\mathcal E(I_{\tau_k-N_{\mathrm{hist}}+1:\tau_k+H})$ and sample $\epsilon\sim\mathcal N(0,I)$. 
At flow noise level $\sigma$, the scheduler forms $Z_{\sigma}=(1-\sigma)Z+\sigma\epsilon$ with target velocity $v^*=\epsilon-Z$.
The first $m\in\{1,3\}$ latent frames encode the history
prefix, with $N_{\mathrm{hist}}=4(m-1)+1$. They are replaced by lightly
perturbed clean latents and assigned timestep zero.
Reference latents
$Z^{\mathrm{ref}}$ are prepended and also assigned timestep zero. VACE injects
control residuals into DiT blocks
$\mathcal L=\{0,4,8,12,16,20,24,28\}$ while leaving reference tokens
unmodified. Define $\widetilde Z_{\sigma}=[Z^{\mathrm{ref}},Z_{\sigma}]$ and
$\widetilde v^*=[0,\epsilon-Z]$. With $M_m$ masking the references and
conditioned prefix, training minimizes only future-latent error:
\begin{equation}
\mathcal L_{\mathrm{FM}}(\psi)=
\mathbb E_{Z,\epsilon,\sigma}\!\left[
w(\sigma)\,
\frac{\left\|(1-M_m)\odot
\left(v_{\psi}(\widetilde Z_{\sigma},\sigma;U_k)-
\widetilde v^*\right)\right\|_2^2}{\lVert1-M_m\rVert_1}
\right].
\label{eq:app-flow-matching-loss}
\end{equation}
We train rank-32 LoRA adapters in Wan attention and feed-forward projections,
$W'=W+(\alpha/r)BA$ with $r=\alpha=32$, together with the configured VACE
blocks and input projection. Control dropout, random VACE strength, and
perturbations of noninitial history latents reduce dependence on perfect
teacher-forced controls and histories.

\subsection{Causal Rollout Implementation}

For the bootstrap chunk, the only real frame and pose are duplicated to satisfy
the two-view minimum of the deployed physical head. Later, the reconstruction
uses only recent and sparse keyframes from the frozen generated prefix, and a
flow-residual mask suppresses dynamic objects before mapping. Genesis resumes
exactly from the simulator checkpoint at each chunk boundary, and every new
video chunk is appended without regenerating its prefix.

\section{Intrinsic Evaluation of PhyMap}
\label{app:phymap}

We evaluate normalized friction, roughness, and stiffness with valid-pixel
weighting. MAE is the primary metric; RMSE emphasizes larger errors. Bias is
the signed mean of prediction minus target and is best near zero. Aggregate
scores average the three properties equally, and bold marks the best result.
Table~\ref{tab:phymap-intrinsic} compares feature and terrain-loss variants.

\begin{table}[!htbp]
\centering
\caption{Feature and terrain-loss ablations.}
\label{tab:phymap-intrinsic}
\resizebox{\linewidth}{!}{%
\begin{tabular}{ccc|ccc|ccc|ccc}
\toprule
\shortstack[c]{DINO\\Feature} &
\shortstack[c]{DPT\\Feature} &
\shortstack[c]{Terrain\\Loss} &
\multicolumn{3}{c|}{\begin{tabular}[c]{@{}ccc@{}}\multicolumn{3}{c}{Friction}\\ MAE$\downarrow$ & RMSE$\downarrow$ & Bias$\to0$\end{tabular}} &
\multicolumn{3}{c|}{\begin{tabular}[c]{@{}ccc@{}}\multicolumn{3}{c}{Roughness}\\ MAE$\downarrow$ & RMSE$\downarrow$ & Bias$\to0$\end{tabular}} &
\multicolumn{3}{c}{\begin{tabular}[c]{@{}ccc@{}}\multicolumn{3}{c}{Stiffness}\\ MAE$\downarrow$ & RMSE$\downarrow$ & Bias$\to0$\end{tabular}} \\
\midrule
$\checkmark$ & $-$ & $-$ & 0.03939 & 0.08407 & -0.01216 & 0.04343 & 0.08378 & 0.00314 & 0.06538 & 0.13945 & -0.00575 \\
$-$ & $\checkmark$ & $-$ & 0.06787 & 0.12288 & -0.01620 & 0.08046 & 0.11739 & 0.00480 & 0.12566 & 0.19969 & 0.00415 \\
$\checkmark$ & $-$ & $\checkmark$ & 0.03706 & 0.08328 & -0.00955 & 0.04122 & 0.08117 & \textbf{0.00114} & 0.06128 & 0.13697 & \textbf{-0.00114} \\
$-$ & $\checkmark$ & $\checkmark$ & 0.06300 & 0.11378 & -0.01350 & 0.07317 & 0.11280 & 0.00224 & 0.11811 & 0.19259 & -0.01469 \\
$\checkmark$ & $\checkmark$ & $-$ & 0.03828 & 0.08472 & -0.01154 & 0.04549 & 0.08386 & 0.00979 & 0.07199 & 0.13925 & -0.01367 \\
$\checkmark$ & $\checkmark$ & $\checkmark$ & \textbf{0.03546} & \textbf{0.08048} & \textbf{-0.00930} & \textbf{0.03845} & \textbf{0.07633} & 0.00318 & \textbf{0.05806} & \textbf{0.12911} & -0.00691 \\
\bottomrule
\end{tabular}}
\end{table}

Fused features with terrain supervision lead on seven of nine metrics and
reduce friction, roughness, and stiffness MAE from 0.03828/0.04549/0.07199 to
0.03546/0.03845/0.05806. DINO with terrain supervision gives the smallest
roughness and stiffness Bias.

\paragraph{Cross-domain generalization.} We train on all domains except $D_k$
and test on $D_k$. In Table~\ref{tab:cross-dataset}, on-robot data gives the
lowest friction and stiffness MAE (0.01798/0.04176), off-road data gives the
lowest roughness MAE (0.02634), and web data has the highest errors.
\begin{table}[!htbp]
\centering
\caption{Cross-domain generalization.}
\label{tab:cross-dataset}
\resizebox{\linewidth}{!}{%
\begin{tabular}{l|ccc|ccc|ccc}
\toprule
Held-out source &
\multicolumn{3}{c|}{\begin{tabular}[c]{@{}ccc@{}}\multicolumn{3}{c}{Friction}\\ MAE$\downarrow$ & RMSE$\downarrow$ & Bias$\to0$\end{tabular}} &
\multicolumn{3}{c|}{\begin{tabular}[c]{@{}ccc@{}}\multicolumn{3}{c}{Roughness}\\ MAE$\downarrow$ & RMSE$\downarrow$ & Bias$\to0$\end{tabular}} &
\multicolumn{3}{c}{\begin{tabular}[c]{@{}ccc@{}}\multicolumn{3}{c}{Stiffness}\\ MAE$\downarrow$ & RMSE$\downarrow$ & Bias$\to0$\end{tabular}} \\
\midrule
Indoor & 0.02579 & 0.05398 & -0.00124 & 0.03789 & 0.07827 & 0.00386 & 0.04900 & 0.11268 & -0.00031 \\
On-robot & 0.01798 & 0.03916 & -0.00214 & 0.03195 & 0.06509 & 0.00422 & 0.04176 & 0.10880 & -0.00875 \\
Off-road & 0.02920 & 0.05527 & -0.00166 & 0.02634 & 0.05961 & -0.00257 & 0.05676 & 0.13614 & 0.00435 \\
Web collection & 0.05587 & 0.11495 & -0.01973 & 0.05027 & 0.09069 & 0.00694 & 0.08374 & 0.16547 & -0.00793 \\
Overall & 0.03419 & 0.07601 & -0.00702 & 0.03776 & 0.07615 & 0.00331 & 0.05994 & 0.13537 & -0.00298 \\
\bottomrule
\end{tabular}}
\end{table}

\paragraph{Comparison with VLM-based physical-property prediction methods.}
We compare \emph{PhyMap} with VLM-based physical-property prediction methods:
PUGS~\citep{shuai2025pugs},
PhysGS~\citep{chopra2026physgs}, SiPhy~\citep{le2026siphy}, and
Qwen3.8~\citep{qwen38} in Table~\ref{tab:physical-map-comparison}.

\begin{table}[!htbp]
\centering
\caption{Comparison with VLM-based physical-property prediction methods.}
\label{tab:physical-map-comparison}
\resizebox{\linewidth}{!}{%
\begin{tabular}{lccccccc}
\toprule
Method & MAE$\downarrow$ & RMSE$\downarrow$ & Bias$\to0$ & Pearson$\uparrow$ & Spearman$\uparrow$ & Acc@0.10 (\%)$\uparrow$ & FPS$\uparrow$ \\
\midrule
PUGS       & 0.1412 & 0.2435 & -0.0710 & 0.5274 & 0.5337 & 66.30 & 0.029 \\
PhysGS     & 0.0883 & 0.1819 & -0.0248 & 0.6258 & 0.5626 & 75.71 & 0.054 \\
SiPhy      & 0.0824 & 0.1785 & -0.0332 & 0.6833 & 0.6899 & 82.21 & 0.029 \\
Qwen3.8    & 0.0799 & 0.1746 & -0.0312 & 0.6935 & 0.6956 & 82.66 & 0.057 \\
\textbf{Ours} & \textbf{0.0459} & \textbf{0.0895} & \textbf{-0.0100} & \textbf{0.9116} & \textbf{0.8489} & \textbf{87.13} & \textbf{6.231} \\
\bottomrule
\end{tabular}}
\end{table}

\begin{figure}[!htbp]
\centering
\includegraphics[width=\linewidth]{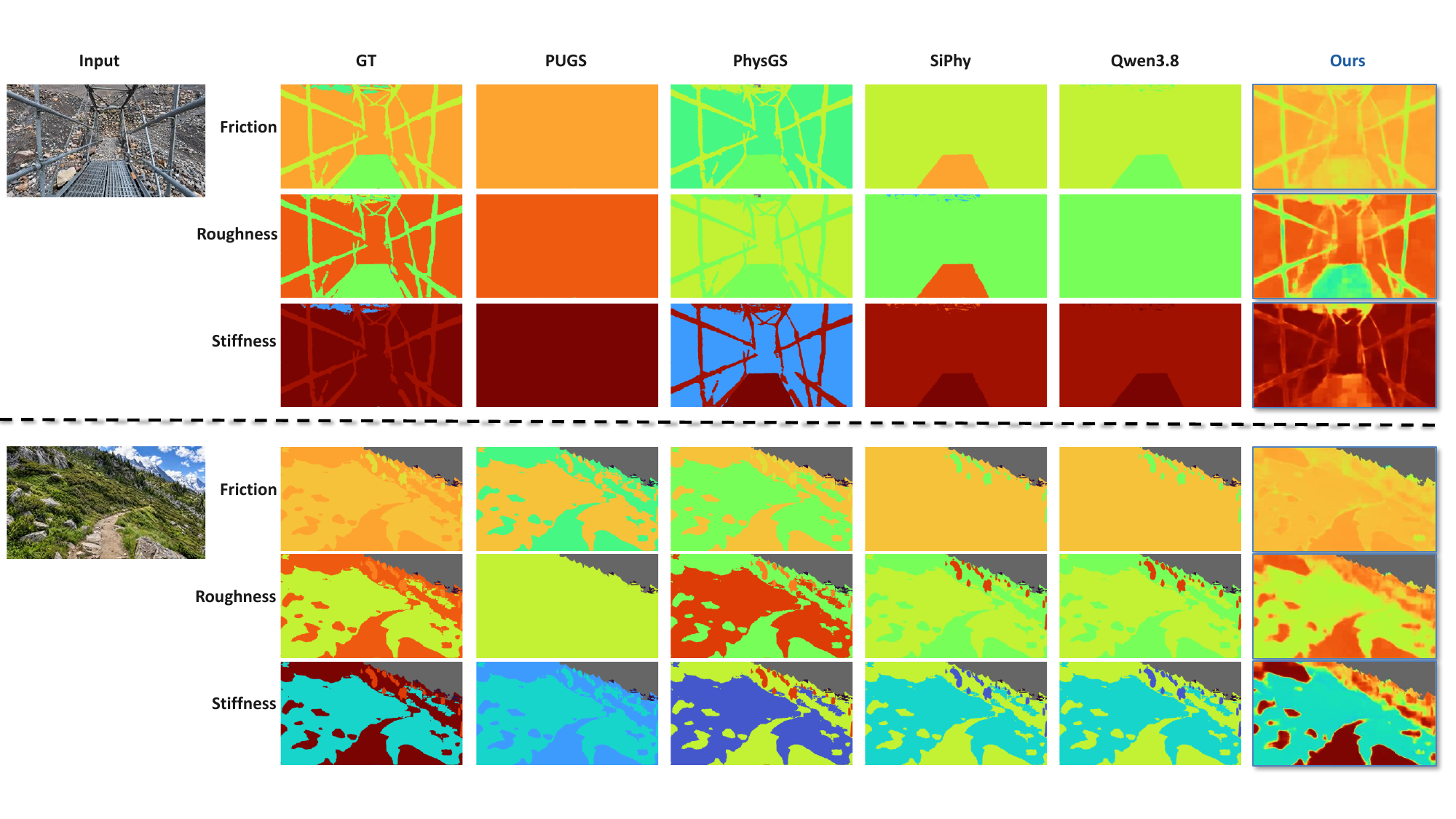}
\caption{Qualitative comparison with VLM-based physical-property prediction
methods.}
\label{fig:physical-map-vlm-comparison}
\end{figure}

\begin{figure}[!htbp]
\centering
\includegraphics[width=\linewidth]{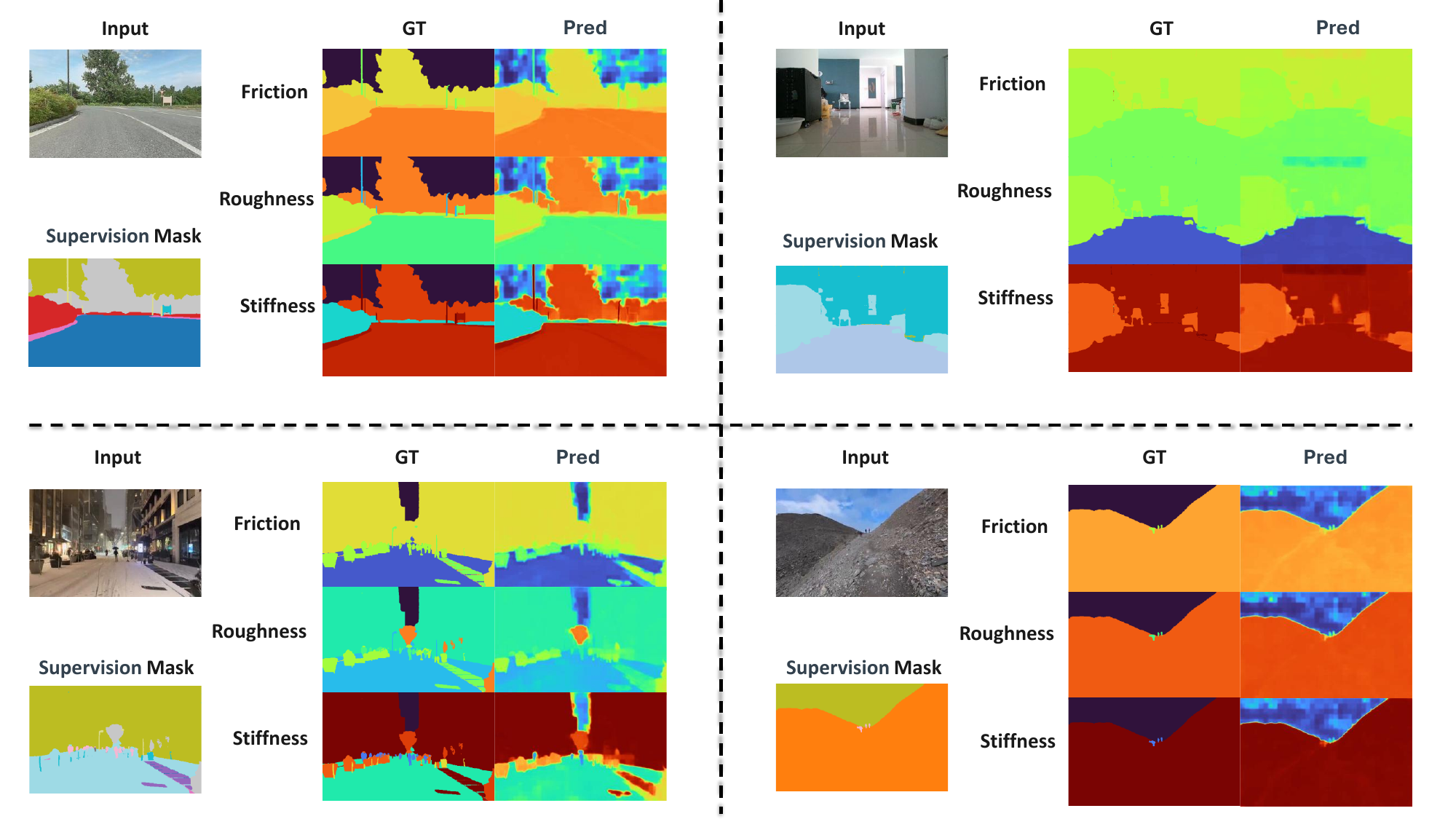}
\caption{Qualitative results of the \emph{Physcene--Image} dataset from four data sources: the on-robot collection,
indoor dataset, web collection, and off-road dataset.}
\label{fig:physical-map-data-source-visualization}
\end{figure}

\emph{PhyMap} leads every accuracy metric and reaches 6.231 FPS versus 0.057 FPS for
the fastest baseline. Baselines use semantic recognition followed by a
class-to-prior lookup, whereas PhyMap predicts the maps directly. The targets
are normalized label-mapped priors.
Figure~\ref{fig:physical-map-vlm-comparison} compares predictions across baselines, showing that \emph{PhyMap} produces maps that more closely match
the reference physical priors .In addition, Figure~\ref{fig:physical-map-data-source-visualization} shows representative examples from the four data sources of \emph{Physcene--Image}: on-robot recordings, public indoor datasets, public off-road datasets, and web videos.

\begin{figure}[!htbp]
\centering
\includegraphics[width=\linewidth]{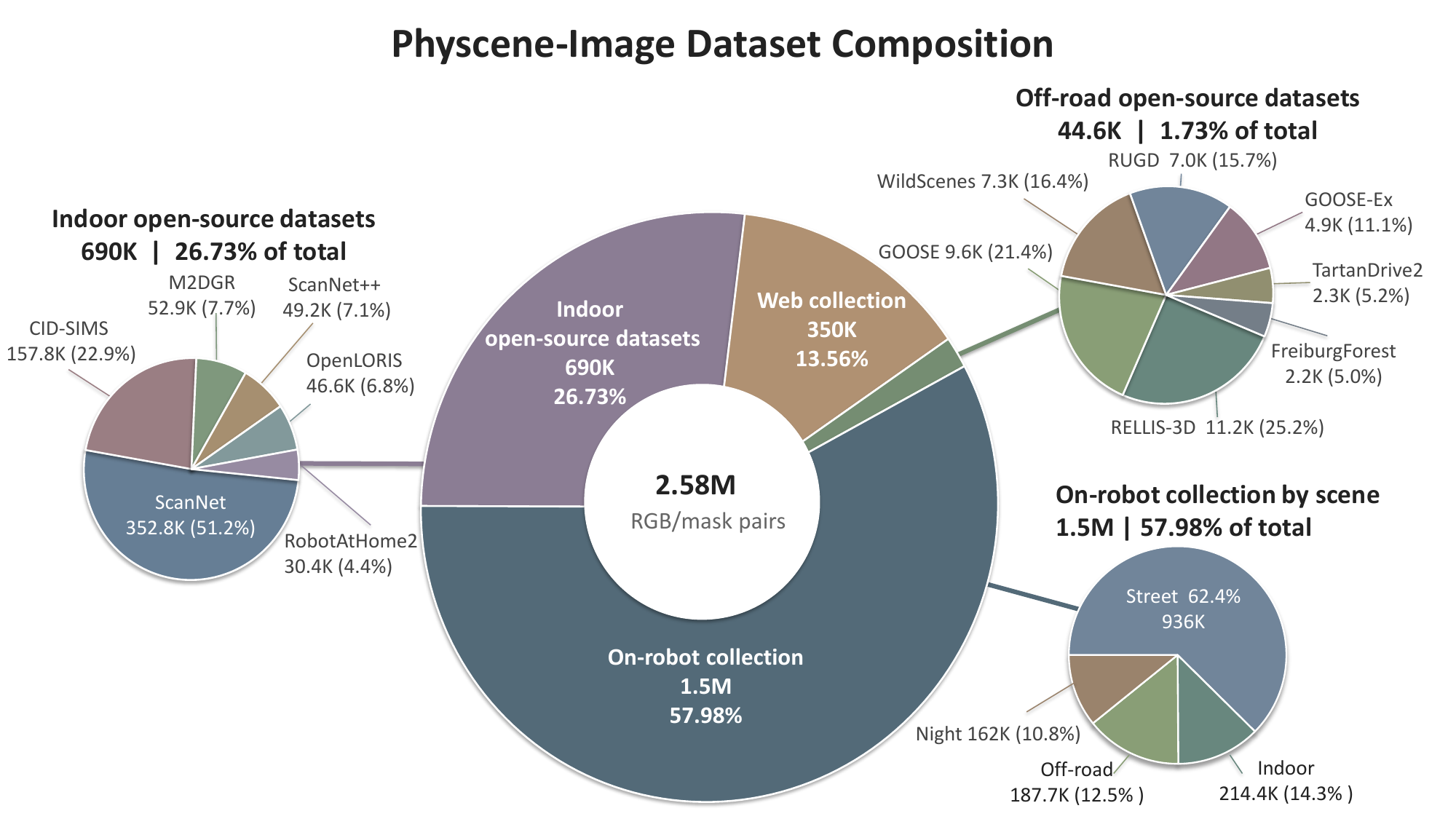}
\caption{Composition of the \emph{Physcene--Image} dataset development.}
\label{fig:phymap-data-distribution}
\end{figure}

% Figure~\ref{fig:phymap-training-data-distribution} summarizes 2.58M unique
% RGB-mask pairs: 1.5M on-robot (57.98\%), 690K indoor open-source (26.73\%),
% 44.6K off-road open-source (1.73\%), and 350K web (13.56\%). The latter three
% sources broaden terrain, appearance, and capture diversity;
% Figure~\ref{fig:physical-map-data-source-visualization} shows representative
% examples from all four sources. All masks are harmonized into Terrain32 and
% mapped to the same normalized physical targets.
\paragraph{Physcene--Image Dataset Composition}
Figure~\ref{fig:phymap-data-distribution} summarizes approximately 2.58M unique RGB--mask pairs: 1.5M on-robot (57.98\%), 690K from public indoor datasets (26.73\%), 44.6K from public off-road datasets (1.73\%), and 350K from web videos (13.56\%).

The on-robot subset is derived from \emph{Physcene--Robot} through frame extraction and deduplication, covering street, off-road, indoor, and nighttime scenes.
Training, validation, and test scenes are strictly disjoint and follow a 7:2:1 split.
The indoor subset combines ScanNet~\citep{dai2017scannet}, CID-SIMS~\citep{zhang2024cidsims}, M2DGR~\citep{yin2021m2dgr}, ScanNet++~\citep{yeshwanth2023scannetpp}, OpenLORIS~\citep{shi2020openloris}, and RobotAtHome2~\citep{ambrosio2023robotathome2}.
The off-road subset includes RELLIS-3D~\citep{jiang2022rellis3ddatasetdatabenchmarks}, GOOSE~\citep{mortimer2024goose}, WildScenes~\citep{vidanapathirana2025wildscenes}, and RUGD~\citep{RUGD2019IROS}.
Additional off-road samples come from GOOSE-Ex~\citep{hagmanns2025gooseex}, TartanDrive2~\citep{sivaprakasam2024tartandrive}, and FreiburgForest~\citep{valada2016deep}.
These public datasets and web videos broaden terrain, appearance, and capture diversity of the \emph{Physcene--Image} dataset.

\section{Additional Executable-Prior Analysis}
\label{app:executable-prior-analysis}

Friction, roughness, and stiffness respectively control Genesis friction,
contact-impedance width, and contact-compliance time. Collision geometry is
identical across variants; the roughness channel affects contact response
while leaving the collision height field fixed. Friction and roughness vary
spatially in all 34 evaluated clips.

\begin{figure}[!htbp]
\centering
\includegraphics[width=\linewidth]{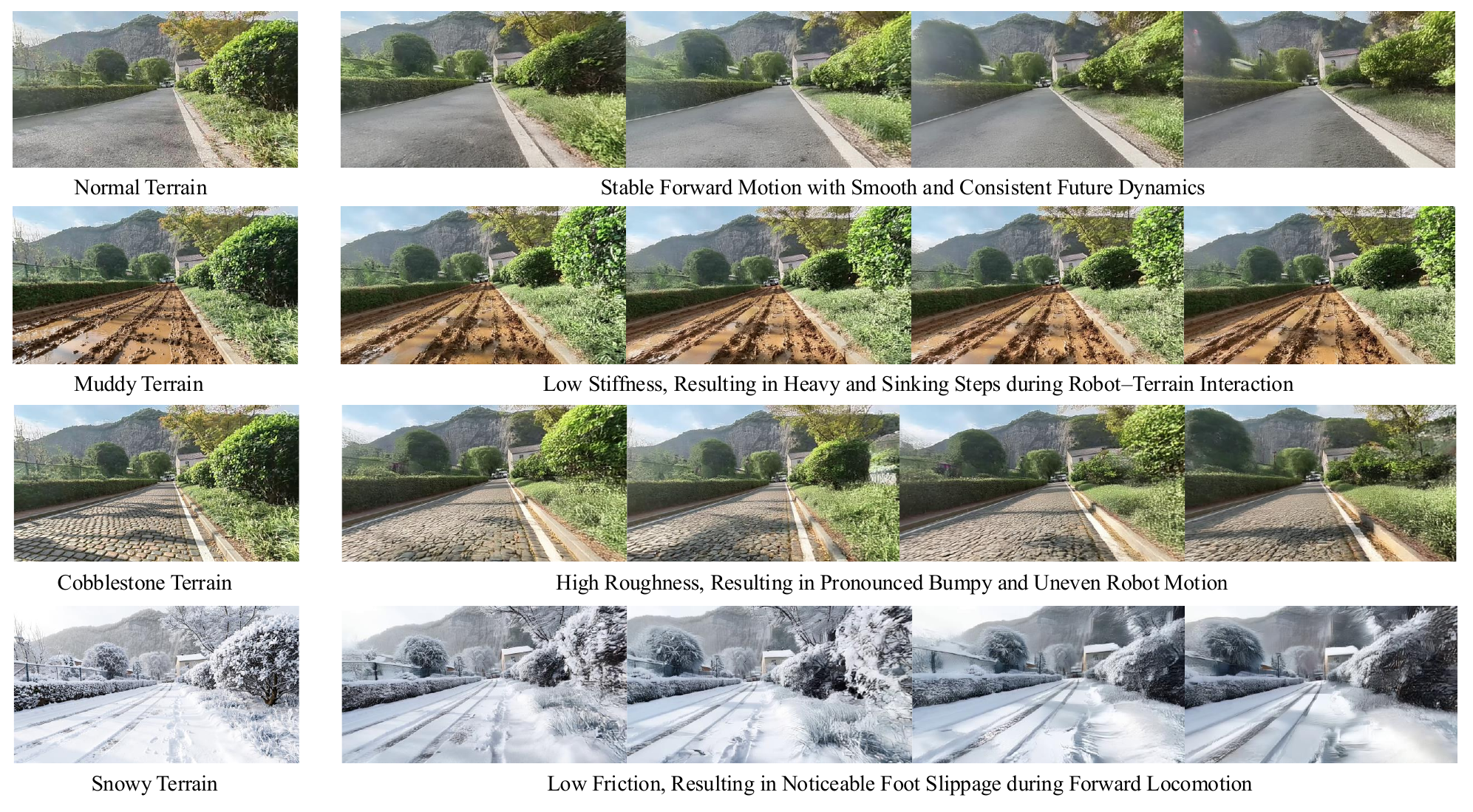}
\caption{Qualitative effects of executable physical priors on robot motion.
Normal terrain produces stable forward motion.
Then low stiffness causes sinking steps on mud, high roughness causes uneven motion
on cobblestone, and low friction causes foot slippage on snow.}
\label{fig:terrain-dynamics}
\end{figure}

The main-paper ablation evaluates the combined physical field. Specifically, the comparison with Uniform \emph{PhyMap} isolates the effect of spatial variation by replacing each channel with its mean over valid terrain. This comparison therefore preserves the mean value of each physical channel, but does not separately quantify the contribution of individual channels or specific material constants. Vertical-motion RMSE is computed between the predicted and recorded vertical displacements, each expressed in its respective initial base frame. Figure~\ref{fig:terrain-dynamics} further illustrates the resulting terrain-dependent dynamics.

\section{Video Generator Ablation}
\label{app:video_ablation}

For our video model, we factorize rendered 3D control and spatial
reference frames following Spatia~\citep{zhao2025spatiavideogenerationupdatable}, then evaluate
action control alone and with both spatial conditions. All variants share the
same checkpoint, prompts, trajectories, causal-history prefixes, and initial
noise; the unconditioned variant still receives the shared prompt and history.

The controlled in-domain evaluation uses \emph{Physcene--Robot} scenes and one fixed seed.
Metrics are computed on generated future frames after excluding the real
history prefix, averaged within scenes and then across scenes. PSNR, SSIM, and
DINO similarity are higher-is-better; the remaining metrics are
lower-is-better. 

\begin{table}[!htbp]
\centering
\begingroup
\setlength{\abovecaptionskip}{2pt}
\setlength{\belowcaptionskip}{3pt}
\caption{Video fidelity across inference-time conditioning variants.
All values are scene-level means computed from generated outputs on their original scales.}
\label{tab:video-ablation}
\setlength{\tabcolsep}{3pt}
\renewcommand{\arraystretch}{0.95}
\resizebox{\linewidth}{!}{%
\begin{tabular}{cccccccccccc}
\toprule
3D control & Spatial refs. & Action control &
PSNR$\uparrow$ & SSIM$\uparrow$ & MAE$\downarrow$ &
Grad. err.$\downarrow$ & Temp. err.$\downarrow$ &
LPIPS$\downarrow$ & DINO Sim.$\uparrow$ & Flow EPE$\downarrow$ & FVD-16$\downarrow$ \\
\midrule
$-$ & $-$ & $-$ & 13.629 & 0.2629 & 0.1572 & 0.0347 & 0.1353 &
0.4062 & 0.7882 & 8.422 & 1153.32 \\
$\checkmark$ & $-$ & $-$ & 15.523 & 0.3062 & 0.1191 & 0.0317 & 0.1321 &
0.2537 & 0.8333 & 6.707 & 572.87 \\
$-$ & $\checkmark$ & $-$ & 13.703 & 0.2639 & 0.1555 & 0.0346 & 0.1353 &
0.3956 & 0.7902 & 8.445 & 1097.23 \\
$\checkmark$ & $\checkmark$ & $-$ &
15.672 & 0.3098 & 0.1167 & 0.0315 & 0.1307 &
0.2465 & 0.8462 & 6.590 & 604.32 \\
\midrule
$-$ & $-$ & $\checkmark$ &
14.091 & 0.2272 & 0.1475 & 0.0321 & 0.1399 &
0.3613 & 0.6005 & 9.844 & 1042.64 \\
$\checkmark$ & $\checkmark$ & $\checkmark$ &
\textbf{16.030} & \textbf{0.3230} & \textbf{0.1106} &
\textbf{0.0314} & \textbf{0.1280} & \textbf{0.2337} &
\textbf{0.8758} & \textbf{6.292} & \textbf{567.39} \\
\bottomrule
\end{tabular}}
\endgroup
\end{table}

Adding 3D control to the reference-only variant improves PSNR by 1.969
[1.144, 2.905] and SSIM by 0.0459 [0.0284, 0.0657]. Spatial references give
smaller gains over control alone (0.149 PSNR and 0.0036 SSIM) and improve most
metrics, although FVD-16 increases from 572.87 to 604.32. Combining action
control with both spatial conditions performs best across all nine metrics. Thus, 3D control contributes most to fidelity,
while the full conditioning combination is strongest overall. This experiment
measures frame fidelity and temporal consistency.

\section{Generated-Observation Physical Remapping}
\label{app:generated_remapping}

We compare four inputs to \emph{PhyMap} at matched timestamps and camera poses:
a single frame, generated visual history, real visual history, and
ground-truth (GT) geometry. The keyframe-selection rule and physical mapping
are fixed before test evaluation. Table~\ref{tab:generated-remapping} reports
geometry error in metres, coverage in percent, physical-property ranking,
Genesis parameter error, and downstream behavior error. Bold marks the better
of the two causal settings, Single Frame and Generated History. The shaded
Real History and GT Geometry rows are reference conditions using future real
observations and ground-truth geometry, respectively.
\begin{table}[!htbp]
\centering
\caption{Physical remapping with different visual inputs.}
\label{tab:generated-remapping}
\resizebox{\linewidth}{!}{%
\begin{tabular}{lccccc}
\toprule
Input to PhyMap & Geometry error$\downarrow$ & Coverage$\uparrow$ & Physics rank$\uparrow$ & Genesis param. error$\downarrow$ & Behavior error$\downarrow$ \\
\midrule
Single Frame & 0.045 & 34.6 & 0.244 & 0.029 & 1.274 \\
\textbf{Generated History} & \textbf{0.039} & \textbf{53.2} &
\textbf{0.351} & \textbf{0.026} & \textbf{1.269} \\
\rowcolor{gray!15} Real History & 0.031 & 54.3 & 1.000 & 0.019 & 1.221 \\
\rowcolor{gray!15} GT Geometry & 0.000 & 100.0 & 0.680 & 0.000 & 1.240 \\
\bottomrule
\end{tabular}}
\end{table}

Compared with a single frame, generated history reduces geometry error from
0.045 to 0.039~m, increases coverage from 34.6\% to 53.2\%, and raises physics
rank from 0.244 to 0.351. It also reduces Genesis parameter error from 0.029 to
0.026 and downstream behavior error from 1.274 to 1.269. Real History remains
the strongest visual-history reference. GT Geometry provides perfect geometric
coverage but is not an upper bound on physical ranking or downstream behavior.
The causal comparison shows that generated observations improve
geometric coverage and physical ranking for the next map update; the reduction
in downstream behavior error is smaller, at 0.005.

\section{Causality Audit}
\label{app:causality_failure}

\paragraph{Trajectory-conditioned protocol.} For every rollout chunk, we record visual
inputs, history/reference indices, geometry and pose sources, file hashes,
causal cutoffs, and cache dependencies. For the non-oracle causal variants,
only real RGB frame~0 initializes the rollout, and future real RGB is used
only as evaluation ground truth. Recorded camera poses condition every method,
and Genesis is not executed. This protocol evaluates visual-history and
geometry updates under an observed trajectory. The evaluation with
Genesis-predicted motion is reported separately in
Appendix~\ref{app:strict_genesis_rollout}.

\metrics Map drift is symmetric Chamfer RMSE in shared world coordinates,
measured in metres. Video error is mean $1-\mathrm{SSIM}$ over one-second
windows ending at 4.8 and 9.6~s. The paired Full-GT map supplies the reference,
so its map drift is zero by construction. In Table~\ref{tab:causal-rollout},
bold marks the best score among non-oracle methods, and shaded rows denote
oracle reference conditions.

Table~\ref{tab:causal-rollout} compares static Frame-0 conditioning, generated
visual history, geometry updates, their combination, and two oracle reference
conditions over 50 scenes. The full model achieves the lowest mean map drift
and video error among the four non-oracle methods at both horizons. Relative
to Static Frame-0, it reduces map drift by 27.2\% at 4.8~s and 42.3\% at
9.6~s. Its video error is 0.661 and 0.662 at these horizons, respectively,
compared with 0.675 and 0.682 for Geometry Update Only. These results support
combining generated visual history with online geometry updates to improve
map consistency and video fidelity under observed camera motion.

Both oracle conditions achieve lower video error than the full model at both
horizons. GT-History yields 0.653 at 4.8~s and 0.652 at 9.6~s, while Full-GT
yields 0.652 and 0.649, respectively. The full model therefore retains a gap
to both oracle references despite its advantage over the other non-oracle
configurations.
\begin{table}[!htbp]
\centering
\caption{Trajectory-conditioned causal rollout on 50 scenes at 4.8 and 9.6~s.}
\label{tab:causal-rollout}
\resizebox{\linewidth}{!}{%
\begin{tabular}{lcccc}
\toprule
Method & Map drift @4.8s$\downarrow$ & Map drift @9.6s$\downarrow$ & Video error @4.8s$\downarrow$ & Video error @9.6s$\downarrow$ \\
\midrule
Static Frame-0 & 2.093 & 2.949 & 0.683 & 0.673 \\
Visual History Only & 2.092 & 2.952 & 0.700 & 0.703 \\
Geometry Update Only & 1.535 & 1.778 & 0.675 & 0.682 \\
\textbf{Full model (observed poses)} & \textbf{1.523} & \textbf{1.702} & \textbf{0.661} & \textbf{0.662} \\
\rowcolor{gray!15} GT-History Oracle & 0.146 & 0.189 & 0.653 & 0.652 \\
\rowcolor{gray!15} Full-GT Oracle (observed poses) & 0.000 & 0.000 & 0.652 & 0.649 \\
\bottomrule
\end{tabular}}
\end{table}

\section{Strict Genesis-in-the-Loop Long-Horizon Rollout}
\label{app:strict_genesis_rollout}

We evaluate whether updating executable geometry and physical fields from
generated observations improves motion prediction over successive Genesis
rollouts. Inference starts from the initial real RGB frame and causal robot
state and follows a recorded control sequence.

\protocol For every non-oracle rollout, inference receives exactly the first
real RGB frame $I_0$, the latest causal robot state $s_0$ at the initial time,
the raw recorded control sequence, and fixed camera, robot, controller, and
calibration assets. Future real RGB, LiDAR, robot state, and camera pose are
isolated behind the evaluation interface and cannot affect mapping, reference
selection, simulation, video generation, or termination. When a two-view
initialization is required, the first frame is duplicated rather than reading a
future observation.

Each rollout spans four consecutive 2.4~s chunks (9.6~s total). At chunk $k$,
PhyMap reconstructs geometry and physical fields from $I_0$ and the method's
own generated history; the updated executable terrain is instantiated in
Genesis; Genesis continues the recorded controls from the previous predicted
state; the fixed camera extrinsics convert simulated robot states into a camera
trajectory; and our trained video prediction model generates the next
video chunk. The generated frames become the only new visual observations
available to chunk $k+1$. Thus the endpoint follows three completed
generated-observation map updates followed by further Genesis simulation.
Robot pose and velocity, joint state, controller memory and targets, camera
follow state, command-filter state, simulator time, and random seeds are
serialized across chunk boundaries. Simulation resumes from the predicted
state without resets to future ground truth. Evaluation uses initial-frame
alignment, with no trajectory-wide alignment or fitted scale correction.

\datasplit The evaluation uses 50 independent \emph{Physcene--Robot} scenes with
synchronized command and state streams and three fixed generation seeds per
scene. Scene identities, temporal windows, seeds, checkpoints, calibration,
and failure handling were frozen before inspecting model outcomes. Comparison
of the frozen split manifests verified that the scenes are disjoint from all
training, calibration, controller-tuning, and hyperparameter-selection data.
Table~\ref{tab:strict-genesis-rollout} reports the held-out results.

\baselines All variants share the initial observation, robot state, controls,
controller, video model, sampling configuration, valid-terrain mask, unknown
region completion rule, and paired random seeds. \emph{Frozen Initial Map}
keeps the executable map inferred from $I_0$ fixed. \emph{Online Geometry +
Frozen Physics} updates geometry from generated history while retaining the
initial physical field. \emph{Online Geometry Only} updates geometry but uses
fixed uniform physical parameters. \emph{Online Uniform PhyMap} updates both
geometry and physical predictions but broadcasts each channel's valid-terrain
mean. \emph{Full DR-WM} retains the complete updated spatial physical field.
The primary controlled comparisons are Full DR-WM against Frozen Initial Map,
Online Geometry + Frozen Physics, and Online Uniform PhyMap.

\metrics Base ATE is computed directly from the Genesis base trajectory, and
camera rotation error is computed from the camera pose induced by the simulated
robot state and fixed rig extrinsics. We additionally report velocity RMSE.
All pose metrics use timestamp synchronization and initial-frame alignment
only. Table~\ref{tab:strict-genesis-rollout} reports ATE at 4.8 and 9.6~s,
camera rotation error at 9.6~s, and velocity RMSE at 9.6~s. Computational
failures and physical events are retained as separate outcome categories.
Metrics are first
averaged over the three seeds within each scene and then averaged over scenes.
Bold marks the lowest error in each column.

\begin{table}[!htbp]
\centering
\caption{Strict Genesis-in-the-loop motion prediction on 50 held-out scenes.}
\label{tab:strict-genesis-rollout}
\resizebox{\linewidth}{!}{%
\begin{tabular}{lcccc}
\toprule
Method & ATE @4.8s (m)$\downarrow$ & ATE @9.6s (m)$\downarrow$ & RotErr @9.6s ($^\circ$)$\downarrow$ & Vel. RMSE @9.6s (m/s)$\downarrow$ \\
\midrule
A. Frozen Initial Map & 0.617 & 1.345 & 16.307 & 0.534 \\
B. Online Geo. + Frozen Physics & 0.608 & 1.334 & 13.773 & 0.425 \\
C. Online Geometry Only & 0.653 & 1.404 & 15.781 & 0.392 \\
D. Online Uniform PhyMap & 0.604 & 1.267 & 14.198 & 0.359 \\
\textbf{E. Full DR-WM} & \textbf{0.568} & \textbf{1.242} & \textbf{13.096} & \textbf{0.322} \\
\bottomrule
\end{tabular}}
\end{table}

\paragraph{Results.} Full DR-WM achieves the lowest error in all four columns.
Compared with Frozen Initial Map, it reduces ATE from 0.617 to 0.568~m at
4.8~s and from 1.345 to 1.242~m at 9.6~s. At 9.6~s, camera rotation error
decreases from 16.307$^\circ$ to 13.096$^\circ$, and velocity RMSE decreases
from 0.534 to 0.322~m/s. These correspond to reductions of 7.7\%, 19.7\%,
and 39.7\% in ATE, rotation error, and velocity RMSE at 9.6~s, respectively.

Updating the full physical field also improves on Online Geometry + Frozen
Physics: ATE at 9.6~s decreases from 1.334 to 1.242~m, and velocity RMSE
decreases from 0.425 to 0.322~m/s. Compared with Online Uniform \emph{PhyMap},
retaining spatial variation reduces ATE at 9.6~s from 1.267 to 1.242~m and
velocity RMSE from 0.359 to 0.322~m/s. The smaller ATE difference in this
comparison indicates a modest additional trajectory benefit from spatial
variation, alongside a larger relative improvement in velocity prediction.
Online Geometry Only has higher ATE than Frozen Initial Map at both horizons,
despite lower velocity RMSE. 
Together, the reported means support updating geometry and spatial physical fields jointly during generated-observation rollouts.
On independent test scenes, this supports improved subsequent Genesis motion
prediction from generated-observation map updates under recorded controls.

\end{document}